\pdfoutput=1

\documentclass[11pt]{article}

\usepackage[final]{acl}

\usepackage{times}
\usepackage{latexsym}

\usepackage{graphicx}
\usepackage{multirow}
\usepackage{hyperref}
\usepackage{booktabs}
\usepackage[T1]{fontenc}
\usepackage{rotating}
\usepackage{pgfplotstable}
\pgfkeys{/pgf/number format/.cd,1000 sep={}}
\pgfplotsset{compat=1.17}
\usetikzlibrary{patterns}
\usepackage[normalem]{ulem}
\usepackage{natbib}
\usepackage{adjustbox}
\usepackage{amsmath}
\usepackage{pifont}
\usepackage{pdflscape}
\usepackage{fancybox}
\usepackage{amssymb}
\usepackage{tikz}
\usetikzlibrary{shapes.geometric, arrows}

\usepackage[utf8]{inputenc}

\usepackage{microtype}
\usepackage{inconsolata}

\useunder{\uline}{\ul}{}

\title{LLMs with Industrial Lens: Deciphering the Challenges and Prospects -- A Survey}

\author{Ashok Urlana\textsuperscript{1} \enspace \enspace Charaka Vinayak Kumar\textsuperscript{1} \enspace \enspace  Ajeet Kumar Singh\textsuperscript{1} \\ \enspace \enspace \textbf{Bala Mallikarjunarao Garlapati\textsuperscript{1}} \enspace \enspace \textbf{Srinivasa Rao Chalamala\textsuperscript{1}} \enspace \enspace \textbf{Rahul Mishra\textsuperscript{2}}\\
TCS Research, Hyderabad, India\textsuperscript{1} \enspace \enspace \enspace \enspace \enspace \enspace
IIIT Hyderabad\textsuperscript{2}\\
{\tt ashok.urlana@tcs.com}, {\tt charaka.v@tcs.com}, {\tt ajeetk.singh1@tcs.com} \\{\tt balamallikarjuna.g@tcs.com}, {\tt chalamala.srao@tcs.com}, {\tt rahul.mishra@iiit.ac.in}
}

\begin{document}
\maketitle
\begin{abstract}

Large language models (LLMs) have become the secret ingredient driving numerous industrial applications, showcasing their remarkable versatility across a diverse spectrum of tasks. From natural language processing and sentiment analysis to content generation and personalized recommendations, their unparalleled adaptability has facilitated widespread adoption across industries. This transformative shift driven by LLMs underscores the need to explore the underlying associated challenges and avenues for enhancement in their utilization. In this paper, our objective is to unravel and evaluate the obstacles and opportunities inherent in leveraging LLMs within an industrial context. To this end, we conduct a survey involving a group of industry practitioners, develop four research questions derived from the insights gathered, and examine 68 industry papers to address these questions and derive meaningful conclusions. We maintain the Github\footnote{\url{https://github.com/vinayakcse/IndustrialLLMsPapers}} repository with the most recent papers in the field.  
\end{abstract}


\section{Introduction}
 Large language models (LLMs) have gained significant attention recently for their exceptional performance in various predictive and generation tasks.  Despite their robust performance across various applications, there remains a notable absence of standardized studies focused on the practical utilization of LLMs in industrial applications as well as the corresponding industrial requirements and deployment challenges.

 To comprehend the current state of LLMs adaption in the industry, we propose a unique two-stage approach as follows, 1) we conduct a case study tailored for industry practitioners of LLMs 2) From the case study, we frame several research questions and address them by conducting a survey of industrial research works. 
 
 To conduct the case study, we requested industry practitioners to respond to a questionnaire enquiring about the application domain, corresponding task, and their associated challenges. The detailed analysis of the industrial case study can be found in Appendix~\ref{sec:industry_survey}. We have obtained 26 responses from several mid-sized companies. With this study, we observed that LLM adoptions in the industry are facing critical challenges such as compute requirements, privacy concerns, and open access. Considering these observations, we framed the following research questions to be addressed in this work.  

 \textbf{RQ1.} How do LLMs assist industrial applications, and in what ways? \textbf{RQ2.} What are the primary applications that industries are focusing on, including the associated datasets and evaluation metrics? \textbf{RQ3.} What are the deployment challenges, if any? \textbf{RQ4.} What are the potential directions to maximize the utility of LLMs in industrial applications? 
 \input{plot_application_life_cycle}
 \usetikzlibrary{arrows,shapes,positioning,shadows,trees}
\tikzset{
  basic/.style  = {draw, text width=1cm, drop shadow, font=\small, rectangle, thin, rounded corners=5pt},
  root/.style   = {rectangle, rounded corners, minimum width=2cm, minimum height=0.3cm, text centered, draw=magenta, fill=white!30, thin, rounded corners=5pt},
  level 2/.style = {rectangle,xshift=35pt,  minimum width=2cm, minimum height=0.3cm, text centered, draw=magenta, fill=white!30, thin, rounded corners=5pt},
  level 3/.style = {rectangle, minimum width=2cm, minimum height=0.3cm, thin, fill=white!60, text width=3cm, text centered, draw=magenta, rounded corners=5pt}
}

\begin{figure*}
\centering\small

\begin{tikzpicture}[font=\small,
  level 1/.style={sibling distance=42mm}, edge from parent/.style={->,draw}, >=latex]
\node[root, minimum width=1cm ] { Industrial LLMs Survey}
            [style=edge from parent fork down]
  child {node[level 2] (c1) {Datasets (\S \ref{sec: section_3}) }} 
  child {node[level 2] (c2) {LLMs applications (\S \ref{sec: section_4})}} 
  child {node[level 2] (c3) {Evaluation methods (\S \ref{sec: section_5})}} 
  child {node[level 2] (c4) {Deployment (\S \ref{sec: section_6}) \& Discussions (\S \ref{sec: section_7})}};

    \begin{scope}[every node/.style={level 3}];
    \node [below of = c1, xshift=35pt, yshift=0.35cm, text width=3.5cm] (c11) {Standard datasets (\S \ref{sec: section_3.1})};
    \node [below of = c11, yshift=0.35cm, text width=3.5cm] (c12) {Application specific (\S \ref{sec: section_3.2})};
    \node [below of = c12, yshift=0.20cm, text width=3.5cm] (c13) {Data generation using LLMs (\S \ref{sec: section_3.3})};

\node [below of = c2, xshift=35pt, text width = 3.5cm, yshift=0.35cm] (c21) {Standard NLP tasks (\S \ref{sec: section_4.1})};
\node [below of = c21,  yshift=0.35cm, text width = 3.5cm] (c22) {Code generation (\S \ref{sec: section_4.3})};
\node [below of = c22, yshift=0.35cm, text width = 3.5cm] (c23) {Societal impact (\S \ref{sec: section_4.7})};
\node [below of = c23, xshift = 0.3cm, yshift=0.35cm, text width=4cm] (c24) {Tools \& Frameworks (\S \ref{sec: section_4.2}) };
\node [below of = c24, xshift = 0.5cm, yshift=0.35cm, text width=5cm] (c25) {Retrieval \& Recommendation (\S \ref{sec: section_4.5})};
\node [below of = c25, xshift = 0.5cm, yshift=0.35cm, text width=6cm] (c26) {Trustworthy AI (\S \ref{sec: section_4.4}) and Security (\S \ref{sec: section_4.6})};
\node [below of = c26, xshift = -1.5cm, yshift=0.35cm] (c27) {Miscellaneous (\S \ref{sec: section_4.8})};

\node [below of = c3, xshift=57pt, minimum width=2cm, text width = 5cm, yshift=0.35cm ] (c31) {Lexical (\S \ref{sec: section_5.1})\& semantic-based (\S \ref{sec: section_5.2})};
\node [below of = c31, yshift=0.35cm,  text width = 5cm] (c32) {Application specific metrics (\S \ref{sec: section_5.3})};
\node [below of = c32, yshift=0.35cm, text width = 5cm] (c33) {LLM based metrics (\S \ref{sec: section_5.4})};


\end{scope}

\foreach \value in {1,2,3}
  \draw[->] (c1.195) |- (c1\value.west);

\foreach \value in {1,...,7}
  \draw[->] (c2.195) |- (c2\value.west);

\foreach \value in {1,...,3}
  \draw[->] (c3.195) |- (c3\value.west);


\end{tikzpicture}
\caption{Structure of the paper.}
\label{tab: paper_structure}
\end{figure*}
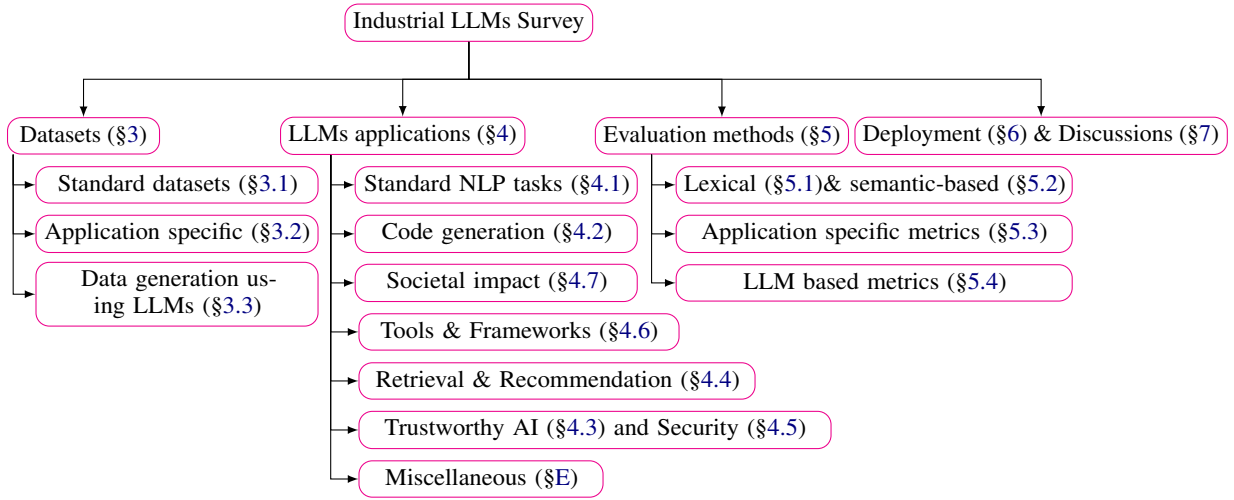 
 With this study, our objective is to answer the above-mentioned research questions. To this end, we collect and analyze 68 research papers specific to industrial applications (see Appendix~\ref{sec:papers_selection}). Further, we categorize the selected papers into 8 broad application domains as shown in Figure~\ref{fig: application_stats}. We observe an uneven distribution of papers related to these application domains. Mostly LLMs are utilized for standard NLP tasks \cite{paper_004, paper_003, paper_075}, code generation tasks \cite{paper_091, paper_093}, building tools and frameworks \cite{paper_068}. However, there is less focus given to security \cite{paper_017} and societal impact \cite{paper_026} related applications.  

 Some of the prominent use cases include the utilization of GPT4 \cite{achiam2023gpt} for automatic summary generation of organizational meetings \cite{paper_004} and the adaption of PaLM \cite{chowdhery2023palm} for low-resource machine translation tasks \cite{paper_075}. Moreover, LLMs are widely used to ensure the security and privacy of the users' data \cite{kim-etal-2021-changes}, mitigate bias, and improve fairness \cite{paper_016}. As depicted in Figure~\ref{tab: paper_structure}, our survey includes the analysis of various datasets, models, associated challenges, and future directions to utilize LLMs for industrial applications.

The key contributions of this survey include, 1) To the best of our knowledge, this is the first survey that provides the industrial perspective on the utilization of LLMs for various applications, 2) We conduct a case study involving industrial practitioners to gain insight into the practical applicability of LLMs, 3) We delve into various datasets, methods and evaluation metrics and corresponding limitations relevant to industrial applications, 4) We discuss the deployment challenges and outline the future directions aimed to maximize the utility of LLMs.

\section{Related Surveys}
Recent emergence of LLMs present significant opportunities for the advancements of various industrial applications \cite{hadi2023survey, kar2023unravelling}. Comprehensive surveys have been conducted to enhance our understanding of the efficacy of LLMs in diverse domains and tasks, such as medical \cite{thirunavukarasu2023large}, finance \cite{paper_051} and reasoning tasks \cite{huang-chang-2023-towards, qiao-etal-2023-reasoning}.

In the realm of code generation, LLMs have gained widespread usage, and a recent survey \cite{zan-etal-2023-large} delves into the associated challenges including code generation and code completion. Few surveys focus on the trust aspects of LLMs \cite{liu2023trustworthy, ishihara-2023-training}, while others offer a detailed exploration of in-context learning techniques \cite{li-2023-practical}, feedback learning models \cite{kirk-etal-2023-past}, and the capability of LLMs to adapt to the constantly evolving world knowledge \cite{zhang-etal-2023-large}. In contrast to existing surveys, our work specifically concentrates on utilization of LLMs for industrial applications. Additionally, this survey provides a comprehensive overview of datasets, models, evaluation metrics, corresponding challenges, and future perspectives.
\section{Benchmark Datasets}
\label{sec: section_3}
This section outlines the datasets used for the development of industrial applications and their creation approaches. These datasets can be broadly divided into two categories (i) popular task-specific benchmark datasets and, (ii) newly proposed datasets, which are automatically created using LLMs, derived either from standard datasets or manually curated with the help of human annotators. A portion of \textbf{RQ2} is addressed in this section.
\subsection{Standard Datasets}
\label{sec: section_3.1}
Due to the ease of access, many papers in our survey utilize task-specific standard datasets to evaluate the performance of LLMs. More details of the datasets and their corresponding applications can be found in Appendix~\ref{sec:application_specific_datasets_appendix}.

\subsection{Application-specific Datasets}
\label{sec: section_3.2}
To better evaluate the capabilities of LLMs, task-specific datasets are more crucial than generic ones. We have identified such datasets and listed them in Appendix~\ref{sec: application_datasets}. This section describes the task-specific datasets and their creation strategies.\\ \textbf{KOSBI.} \citet{paper_010} introduced a social bias detection dataset for the Korean language by using Hyperclova \cite{kim-etal-2021-changes}. This dataset consists of context-sentence pairs, where the context contains safe-unsafe labeled scenarios mentioning a specified demographic group and the corresponding sentence is the textual content. \textbf{DELPHI.} To mitigate bias and fairness in LLMs, \citet{paper_026} propose the DELPHI dataset. The questions present in the dataset are selected from the `Quora Question Pair' dataset\footnote{\href{https://www.kaggle.com/c/quora-question-pairs}{https://www.kaggle.com/c/quora-question-pairs}}. 
\textbf{WikiBias-Aug.} \citet{paper_028} concentrates on evaluating the extent of gender bias present in LLMs-generated reference letters. WikiBias-Aug dataset was generated by selecting instances from WikiBias \cite{sun-peng-2021-men} and altering the male and female names.
In another study, \textbf{CMExam} \cite{paper_107} dataset was introduced to evaluate the performance of LLMs on open-ended QA task for the Chinese medical data. The raw dataset was collected from the Chinese National Medical Licensing Examination portal. Further, this dataset was refined partially with the help of `GPT-assisted annotation' and reviewed by two medical professionals.
\citet{paper_003} propose \textbf{MSQA}, a novel QA dataset designed for Microsoft products and IT technical problems faced by its customers. This dataset helps address the problem of potential oversight of crucial domain-specific information by LLMs. 
\textbf{FETA.} \cite{paper_014} dataset was introduced to evaluate and improve the performance of LLMs on expert domain tasks. It presents vision learning scenarios such as image-to-text and text-to-image retrieval tasks. FETA dataset was obtained from the IKEA's yearly catalogues\footnote{\href{https://github.com/ivc-yz/SSR?tab=readme-ov-file}{https://github.com/ivc-yz/SSR?tab=readme-ov-file}}. 
Another study, \citet{paper_036} explores the effectiveness of LLMs in network management systems by introducing a benchmark called \textbf{NemoEval}. Each instance in the dataset contains a user query or network management instruction and an answer created by human experts. \citet{paper_002} introduce \textbf{LoTNLG \& F2WTQ} to investigate table-to-text capabilities of LLMs. LoTNLG is based on LOGICNLG \cite{chen-etal-2020-logical} test set, where each entry was labeled by two individual annotators based on the logical reasoning type. 
Dataset F2WTQ is based on WTQ \cite{pasupat-liang-2015-compositional} - a human-annotated QA dataset. For each question in WTQ a corresponding entry in F2WTQ was created, that contains a question and a sentence-long response about a table. 
Furthermore, to resolve the bugs in the code generation and completion tasks, \citet{paper_100} propose \textbf{Buggy-HumanEval and Buggy-FixEval.} datasets. Each instance of these datasets consists of a partial code snippet with bugs, failing test-case, and completed code. `buggy-HumanEval' is a subset of `HumanEval' \cite{chen2021evaluating} dataset containing 1896 buggy programs. Whereas `buggy-FixEval' contains 292 realistic buggy programs submitted to coding contests and was adapted from `FixEval' \cite{mahim2022fixeval} and CodeNet \cite{puri2021codenet} datasets. Similarly, \citet{paper_106} introduces \textbf{MBXP, Multilingual HumanEval, MathQA-X.} datasets for evaluating the LLM's performance on code generation and reasoning over multiple programming languages. These datasets were adapted from MBPP \cite{austin2021program}, HumanEval\cite{chen2021evaluating}, MathQA-python\cite{schubotz2018introducing} by converting them to multiple programming languages using the proposed framework in the \cite{paper_106}.

\subsection{Data generation using LLMs} 
\label{sec: section_3.3}
Few works attempt to generate synthetic datasets by utilizing LLMs. \citet{paper_064} introduced FABRICATOR, an open-source toolkit for generating labeled datasets using LLMs for tasks like text classification, entailment, question-answering, and entity generation through zero-shot and few-shot prompting. 
Similarly, \citet{yin-etal-2023-dynosaur} introduced a novel paradigm for curating instruction-tuning datasets. \citet{paper_049} proposes an approach to improve the accuracy of image generative models by augmenting the input prompts with a layout of the required objects. Multiple variants of GPT models have been used to generate the layouts in 2-D and 3-D image and scene generation tasks.

\section{LLMs for Real-world Applications}
\label{sec: section_4}

This section offers an overview of the diverse approaches employed to use LLMs in industrial applications and also answers (\textbf{RQ1}). To leverage the LLMs for any application, usage of proper prompting strategy is necessary. To elaborate on this, we have included details of prompting strategies in Appendix~\ref{sec: taxonomy_llms}.
\subsection{\textbf{Standard NLP tasks}}
\label{sec: section_4.1}
\textbf{Summarization.} Summarization for business documents plays a crucial role in obtaining the salient aspects of a document. In their work, \citet{paper_004} focuses on summarizing Automatic Speech Recognition (ASR) transcripts of organizational meetings. They evaluate the performance of the open-source (LLaMA-2 \cite{touvron2023llama}) and closed-source (GPT-3.5 \cite{GPT3pt5}, GPT-4 \cite{achiam2023gpt}, PaLM-2 \cite{anil2023palm}) LLMs for summary generation. 
Likewise, to obtain the summaries of product titles of an e-commerce website, \citet{paper_018} presents an instruction fine-tuning approach constrained by the length, number of words and specific phrases. Similarly, \citet{paper_098} presents the summary generation of outage understanding of cloud-based servers. Apart from this, LLMs have also been adopted for generating expository write-ups \cite{paper_103} and movie descriptions \cite{paper_021}. \\
\underline{\textit{Limitations:}} Generalizing summarization methods for domain-specific datasets and reducing factual inaccuracies needs attention.
\\
\textbf{Question-answering. } To enrich domain-specific knowledge of LLMs, \citet{paper_003} collected a cloud domain QA dataset and introduced a novel model interaction paradigm. The method involves pre-training a LM on domain-specific data and passing the user's query to LM to extract domain knowledge. The acquired information and the query are given as inputs to LLM for response generation. This method works \underline{\textbf{\textit{only}}} with English data. \\
To facilitate the automatic workflow generation, \citet{paper_080} proposes a novel approach called Flowmind using GPT. It ensures secure interactions between LLMs and proprietary code or data to protect privacy. To perform the visual QA task, \citet{paper_102} proposed the PROMPTCAP - a novel prompt-guided image captioning technique, that takes the prompt to describe the visual entities present in the image in the generated caption. PROMPTCAP is trained on synthetic examples generated by GPT3 and is \underline{\textbf{\textit{limited}}} to knowledge based visual QA tasks.\\ 
\textbf{Machine translation.}  To obtain effective translations for low-resource languages, \citet{paper_075} perform the few-shot prompting on LLaMA and translate English semantic parsing datasets into 50 more languages. However the proposed approach is \underline{\textbf{\textit{constrained}}} by compute resource availability. \\
\textbf{Conversational: } To build a dialog system for public health intervention, \citet{paper_057} designed the CareCall - an open-ended LLM-integrated chatbot to interact with specific individuals and reduce loneliness and emotional distress. CareCall faces challenges that are attributed to \underline{\textbf{\textit{limited}}} personalization. Similarly, \citet{paper_059} employs LLMs to assist augmentative and alternative communication (AAC) users with extended replies and answering biographical questions, aiming to save time and cognitive effort.\\
\textbf{Sentiment analysis. }\citet{paper_099} uses PaLM-540B \cite{chowdhery2023palm} to generate weak sentiment labels for sentiment analysis on financial market data. This approach, training models with weakly labeled data, outperforms traditional methods relying on fully labeled datasets.\\
\textbf{Reasoning. } To assess the reasoning capability of LLMs, \citet{paper_009} conducted experiments on arithmetic problems using GPT-3. Their approach resulted in multiple representations of an arithmetic problem, including algebraic and Pythonic forms. These representations can serve as intermediate steps that may be used for reasoning the solution. In another study, \citet{paper_037} proposes a plug-and-play style framework to improve the reasoning capabilities of LLMs on multi-modal complex tasks by composing external tools and modules depending on the type of task.\\
\citet{paper_051} proposes a novel framework to enhance the steerability of LLMs by introducing `persona', which is based on user characteristics (age, gender). Clusters representing different personas are formed, and their embeddings are appended to input prompts, guiding LLMs through soft prompting. 
In a related domain, \citet{paper_108} explores reasoning on tabular data and presents a unique approach to generate reasoning based on sub-evidences. This strategy decomposes complex tabular data into smaller tables without losing any context. similarly, the complex questions are divided into simple SQL sub-queries. After obtaining answers for the sub-queries using smaller tables as evidence, a final reasoning answer is derived. The performance of proposed approach \underline{\textbf{\textit{degrades}}} with increased context length. \\
\textbf{Table-to-text generation. }This task focuses on creating coherent and contextually accurate descriptions from structured tabular data. 
Building on this concept, \citet{paper_002} conducted a study to investigate the table-to-text capabilities of LLMs across diverse datasets and confirm the potential of LLMs to accurately interpret and convert structured data into descriptive text. Whereas, \citet{paper_032} addressed challenges faced by LLMs in processing semi-structured tabular data with inconsistencies. \citet{paper_013} leveraged the LLMs to enhance table metadata and enrich the glossaries, taxonomies, or ontologies. 
\subsection{\textbf{Code Generation}} 
\label{sec: section_4.3}
Given that LLMs are trained on extensive code repositories, experimental studies have been done to evaluate the capabilities of LLMs for multiple tasks including code generation, code completion, program repair, hint generation, grading feedback, pair programming, contextualized explanation, and task synthesis \cite{paper_091}. One such work \cite{paper_051} explores the usage of LLMs for user-driven webpage modification, particularly useful for accessibility reasons or for less technically literate. The source code of the webpage is divided into smaller portions and passed to the LLM along with user-modified prompts. 

\citet{paper_035} proposes CodePlan - a framework to automate the repository level code editing tasks such as package migration (\textit{C\#}) and temporary code edits (Python) across multiple repositories using the GPT-4 model. However, CodePlan \underline{\textbf{\textit{may not be ideal}}} for dynamic languages. Another study utilizes LLMs to enhance the network management experience by generating high-quality code for graph manipulation tasks \cite{paper_036}. The same approach also tackles issues such as explainability, privacy, and scalability in task-specific code generation. On a similar line, \citet{paper_093} introduces an approach called GRACE to predict the code edits to fix the bugs or to add new features using CODEX and CODET5 \cite{chen2021evaluating, bavarian2022new} models. GRACE \underline{\textbf{\textit{may fail}}} when additional context is required apart from associated edits. Similarly, \citet{paper_100} introduced the concept of a buggy code completion problem to generate code suggestions in contexts that potentially contain bugs.  
\citet{paper_106} introduced a framework to generate execution-based test cases for programming languages beyond Python. It converts Python datasets to multiple languages in a scalable manner. The evaluation of this approach indicates that training a single multilingual model outperforms several monolingual ones. On the other hand, to perform the execution-based evaluation, \citet{paper_053} introduced a static evaluation framework specifically for the Python language, that operates by parsing code snippets into Abstract Syntax Trees (ASTs).

\subsection{\textbf{Trustworthy AI}} 
\label{sec: section_4.4}
Nowadays, service providers are emphasizing on trust in AI, focusing on securing and protecting data privacy. To detect the potential privacy violations in LLMs, \citet{paper_056} proposes ProPILE, a tool that enables external users and in-house LLM developers to create custom prompts to identify if private data is being used for LLM training. To mitigate issues such as perpetuated biases, risk of leaking private information, and hallucinations regarding factual data, \citet{paper_069} introduced a suite of fine-tuned LLMs, along with a comprehensive framework for their evaluation, fine-tuning, and deployment. To specifically handle `hallucinations', \citet{paper_019} proposed the INVITE - a framework that acts as a testbed, enabling the generation of invalid questions by intentionally distorting valid facts. \\
In another study, \citet{paper_016} proposed \textit{Finspector}, a tool that allows users to evaluate models for bias and fairness, particularly in sentences exhibiting stereotypes \cite{paper_027}. The proposed tool's effectiveness on decoder-only models is \underline{\textbf{\textit{not tested}}}. Recently, aligned models are designed to encode ethical considerations, and user values, and incorporate safety guardrails. Towards this aim, \citet{paper_066} introduced an open-source toolkit that simplifies adding programmable guardrails to LLM-based conversational systems.

\subsection{\textbf{Retrieval \& Recommendation}} 
\label{sec: section_4.5}
LLMs demonstrate remarkable language generation abilities to perform various information retrieval tasks. Recently, \citet{paper_047} outlines an approach using LLMs to enhance human-human verbal communication by proactively suggesting relevant visuals that reflect the conversation's visual intent. The authors implemented Visual Captions as a Chrome plugin that aims to render live captions and relevant images directly onto the camera video stream for online video conferences.

To improve the retrieval quality, \citet{paper_044} proposed two techniques (i) document expansion - enriches the document representation by appending additional relevant terms, (ii) query expansion - rewriting the query based on relevant feedback. They proposed a method for query disambiguation where a new query is formed by performing few-shot prompting.  \textit{text-davinci-003} was employed to retrieve a pseudo-document which is concatenated with the original query to form a new query for the retrieval task. In this approach, inference is \underline{\textbf{\textit{slowed down}}} by token-by-token auto-regressive decoding strategy.
In another study, \citet{paper_094} explored the utility of a GPT-3.5 to automatically generate queries and query variants from story descriptions. \citet{paper_014} proposed an approach that combines Multiple Instance Learning (MIL) and CLIP \cite{radford2021learning} foundational models for text-to-image and image-to-text retrieval tasks in expert domains.  Especially non-natural image documents, user manuals, patents and infographics were used.

Similarly, \citet{paper_052} proposed a Large Multimodal Model (LMM) for radiology report generation. This LMM comprises a visual encoder (ResNet50 \cite{he2016deep}), that takes an X-ray as input and outputs visual tokens. Subsequently, the visual features embeddings are used as soft prompts to a causal language model (GPT \cite{radford2019language}) and OpenLLaMA-7B \cite{geng2023openllama}).
\citet{paper_042} proposed a generate-then-read pipeline by utilizing InstructGPT \cite{ouyang2022training} to generate contextual documents based on a given query and then perform the retrieval to produce the final answer. However, this approach has \underline{\textbf{\textit{limited}}} ability to update its knowledge state and adapt to new domains. 

Despite the progress in the conventional item-based recommendations for e-Commerce platforms, \citet{paper_048} utilizes language-based preferences as well to provide more consistent recommendations with the help of LLMs. \citet{paper_078} builds a conversational system and integrates it with a voice-enabled robot for hospitable real-time tourism suggestions. In another study, \citet{paper_114} focuses on evaluating learning-to-rank models that utilize user intentions, derived from LLMs through product reviews, to highlight aspects such as education, fun, and durability while browsing on an e-commerce website. 

\subsection{\textbf{Security}} 
\label{sec: section_4.6}
LLMs, accessible via APIs for advanced reasoning tasks, also pose risks of model extraction attacks, leading to unauthorized replication. To this end, \citet{paper_017} proposed a method to embed watermarks in the language model which will act as a backdoor within the embeddings to ensure reliable copyright verification. 

Task specific fine-tuned LLMs often raise significant legal issues, such as violations of licensing agreements and copyright infringement. Addressing this crucial need, \citet{paper_015} proposed an LLM attribution method. This method links fine-tuned models to their original base models, ensuring integrity and responsible use of LLMs. However, this study \underline{\textbf{\textit{lacks}}} the analysis on large-scale finetuned-models. Additionally, significant advancements in cyber threat intelligence have been made with a pre-trained CTI-BERT model  \cite{paper_001} for cybersecurity, enhancing threat intelligence system accuracy.

\subsection{\textbf{Tools and Frameworks}} 
\label{sec: section_4.2} 
LLMs have significantly advanced the development and evaluation of technologies across various fields by providing access to sophisticated frameworks and tools. Notably, \citet{paper_068} introduced RALLE, a framework for developing Retrieval Augmented LLMs to tackle hallucination issues. Other innovative applications include leveraging LLMs for interactive 3D object creation \cite{paper_025} and domain-specific evaluations in radiology \cite{paper_090}, as well as LIDA by \citet{dibia-2023-lida} for automatic visualization generation. Tools like PromptInfuser integrate AI into mockups via LLM prompts \cite{paper_050}, while another tool \cite{paper_072} employs few-shot prompting with LLMs for generating robotic programming instructions. Additionally, a study \cite{paper_007} utilizes GPT for correlating court judgments with hearings, improving legal analysis. \citet{paper_046} explores LLMs for mobile UI conversational interfaces, but \underline{\textbf{\textit{fails}}} to handle irrelevant and hallucinated responses. Conversely, \citet{paper_074} demonstrates the use of LLMs for automatically adapting to new domains with ``own skills'' bootstrapping.

\subsection{\textbf{Societal Impact}}
\label{sec: section_4.7}
The widespread adoption of LLMs must be tempered by addressing potential societal and human risks such as toxicity and bias that can adversely affect society. To mitigate the unsafe societal bias in LLMs, \citet{paper_010} developed a large-scale dataset - KOSBI focusing on Korean language and culture. It employs filtering-based moderation to reduce societal biases in LLM-generated content. \citet{paper_026} examines the interaction of LLMs with controversial issues and proposes methods to enhance their understanding and management of complex societal debates. 
Similarly, \citet{paper_116} introduces DisasterResponseGPT, a framework that generates actionable plans by leveraging LLMs for efficient disaster response strategy creation, that offers a potential solution to transform disaster response operations. The proposed framework is \underline{\textbf{\textit{unable}}} to process image-based inputs. The detailed the analysis of the miscellaneous applications present in Appendix~\ref{sec: section_4.8}.
\section{Evaluation Approaches}
\label{sec: section_5}
This section mainly discusses various types of evaluation metrics and answer \textbf{RQ2} by grouping the evaluation metrics into the following categories.
\subsection{Lexical overlap-based metrics}
\label{sec: section_5.1} 
Due to their simplicity, many works in our study utilize the lexical overlap-based metrics, which measure the alignment score based on n-grams matching between the source and target. \\\textbf{Examples:} BLEU \cite{papineni-etal-2002-bleu}, ROUGE \cite{lin-2004-rouge}, METEOR \cite{denkowski-lavie-2014-meteor}, Accuracy, F1 Score, ExactMatch. 
\subsection{Semantic overlap-based metrics.} 
\label{sec: section_5.2}
Semantic overlap-based metrics exploit contextual embeddings from pretrained language models to assess the similarity between gold and system-generated outputs. \textbf{Examples:} BERTScore \cite{Zhang*2020BERTScore:}, AlignScore \cite{zha-etal-2023-alignscore}. 
\subsection{Application specific metrics} 
\label{sec: section_5.3}
This section describe the task-specific metrics related to each application, where as complete list of metrics details mentioned in Appendix~\ref{sec: evaluation_metrics_appendix} Table~\ref{tab:metrics_info}.\\
\textbf{Table-to-text generation.} (a). \textit{SP-Acc:} verify the correctness by extracting the meaning representation from the generated sentences and executing against the Table and \textit{NLI-Acc} works as a faithfulness classifier by utilizing a pretrained TableBERT \cite{chen-etal-2020-logical} model. (b). \textit{TAPAS-Acc} and \textit{TAPEX-Acc} metrics uses the TAPAS \cite{herzig-etal-2020-tapas} and TAPEX \cite{liu2021tapex} models as backbone, which are fine-tuned on TableFact \cite{chen2019tabfact} dataset.
\textbf{Question-answering.} (a). \textit{Keyword/Span-Hit-Rate} - creates a distinct keyword or span set by extracting keywords from grounded answers and measure the rate of hits in the generated response \cite{paper_003}. (b). \textit{Can-Answer-Rate} - represents the percentage of answerable questions and evaluates the informativeness \cite{paper_003}. 
\textbf{Retrieval and recommendation systems.} To evaluate the performance of LLMs recommendations, metrics used such as Hit Rate, Normalized Discount Cumulative Gain and Mean Reciprocal Rank. (a). \textit{Rank Biased Overlap:} Quantify the consistency in the retrieved search results for various query variants \cite{webber2010similarity}. (b). \textit{Rank Biased Precision:} Measures as the ratio of expected document relevance to the expected number of examined documents \cite{moffat2008rank}.
\textbf{Societal impact.} (a). \textit{Controversy Acknowledgement Rate:} If LLMs response consists of "As an AI language model..." reminds that the response is non-human perspective and conventionally used as an indicator for the controversy acknowledgment rate. (b). \textit{Comprehensiveness Answer Rate:} Measures the LLMs response diversity and extensive knowledge in the presence of real-time scenarios \cite{paper_026, paper_094}.
\textbf{Code generation.} (a). \textit{Block metrics: } aims to identify the code blocks in need of modifications. (b). \textit{Edit metrics: } measures the correctness of the code modifications \cite{paper_035}. (c). \textit{pass@k:}  require k generations to address a given problem, deeming the problem resolved if any generation successfully passes the corresponding unit tests \cite{chen2021evaluating}.
\textbf{Trustworthy AI.} \textit{Word Embedding Association Test (WEAT): } assess and quantify the associations between different sets of words in word embeddings  \cite{caliskan2017semantics}.
\subsection{LLM Based Evaluation}
\label{sec: section_5.4}
A recent study \cite{paper_003} employed LLM as an evaluator for ranking the responses based on the similarity with the grounded answer.
Due to space constraint, we include the human evaluation details in Appendix~\ref{sec: evaluation_metrics_appendix} Section~\ref{sec:human_eval_appendix}.



 \section{Deployment Challenges}
 \label{sec: section_6}
In this section, we discuss some of the potential deployment challenges and solutions, observed in our study (\textbf{RQ3}).\\
\textbf{Potential privacy risks.} Deploying LLMs on cloud platforms necessitates strong data privacy and security protocols to manage large, sensitive datasets. The main challenges include preventing data breaches and the unauthorized extraction of sensitive information. Implementing robust security and privacy technologies is essential to protect user data while ensuring LLMs remain functional. \\ 
\textbf{Infrastructure.} Infrastructure is crucial for deploying LLMs, influencing factors like processing speed, latency, cost, and training needs. The necessity for high-performance hardware to boost speed and reduce latency enhances user experience but comes at a high cost, demanding careful budgeting. Achieving an optimal balance between cost and performance is crucial for the efficient training and scalability of LLM applications. \\
\textbf{Control the level of AI proactivity.} LLMs should reduce social awkwardness, improve expressiveness, and adapt to different scenarios  \citet{paper_047}. Moreover, LLMs' open-ended nature makes it difficult to customize dialog systems in public health intervention applications \cite{paper_057}.\\
\textbf{Regulations.} GDPR in Europe and CCPA in California introduce major hurdles for deploying LLMs by enforcing strict data handling and intellectual property rules, requiring transparency and fairness. These laws necessitate navigating diverse regulations, complicating LLM deployment. As highlighted by \citet{reg_med}, adhering to these laws in sensitive domains like healthcare is crucial to avoid harm and protect privacy. Therefore, developing a comprehensive framework for LLM compliance is essential for responsible use and interaction with users.

\section{Challenges and Future Perspectives}
\label{sec: section_7}
In this section, we discuss the various challenges and future perspectives to maximize the utility of LLMs for industrial applications (\textbf{RQ4}). 

Maximizing LLMs' potential in healthcare, retail, and finance requires high-quality, labeled datasets. Additionally, with most LLMs being limited to English, \underline{there is significant room} for creating robust multilingual models.

LLMs face challenges in emotional support and long-term memory, impacting conversational applications \cite{paper_057}. In sectors like legal, finance, and healthcare, blending LLMs with \underline{human feedback is crucial} to lowering false positives, underscoring the importance of human oversight in safety-critical applications \cite{paper_090}. Moreover, our analysis, detailed in Appendix~\ref{sec:human_eval_appendix}, reveals that \underline{less than 15\% of studies} conduct human evaluations to assess LLM outputs, indicating a need for more rigorous validation methods.

Despite the adoption of LLMs in applications like disaster response management \cite{paper_116}, public health intervention \cite{paper_057}, and assisting AAC users \cite{paper_059}, there is a noticeable \underline{lack of focus on privacy and security} aspects. Moreover, it is imperative to address the potential risks associated with deploying LLMs in high-stakes scenarios. 

With the \underline{rising demand for multimodal} applications, only a few studies focused on utilizing LLMs for such industrial applications \cite{paper_049, paper_052}. Likewise, more \underline{efforts are needed} on integrating LLMs with voice assistants and Robotics \cite{paper_078}. As shown in Figure~\ref{fig: industrial_utility}, we found that more than 70\% of LLM-based projects are \underline{still in the conceptual phase}. 

\section{Conclusions}
 This survey delves into the utilization of Large Language Models (LLMs) through an industrial lens, with a specific focus on identifying adoption challenges and avenues for promising use cases. Additionally, it meticulously examines the datasets, methodologies, and evaluation metrics used, along with their limitations across different industrial applications of LLMs. By addressing deployment challenges and suggesting future directions, the survey offers a strategic roadmap for industries to optimize the effectiveness of LLMs in their operations. 
\section{Limitations}
With the rapid development of LLMs and the voluminous research in this field, it's not feasible to comprehensively cover all the papers. Recognizing this, our survey has focused specifically on industry-related papers. This allowed us to delve deeper and gain understanding of the unique requirements and challenges faced within industrial applications of LLMs. Due to the page limit, we cannot include all technical details. Retrieval-augmented generation (RAG) has enhanced the performance of the LLMs by injecting external knowledge beyond their static training data.  However, this survey specifically examines LLMs' industry applications, with minimal focus on RAG to focus solely on their direct applications and inherent challenges.
\section{Ethics Statement}
To our knowledge, this study presents minimal ethical concerns. However, to maintain transparency, we provide a detailed analysis of all 68 papers present in the survey in Appendix~\ref{sec:check_list_survey}. Each paper is reviewed by at least three individuals to validate its claims and findings. Furthermore, we did not collect any personal information as part our case study.
\bibliography{custom}
\bibliographystyle{acl_natbib}
\newpage
\appendix
\section{Survey Papers Selection Criteria}
\label{sec:papers_selection}
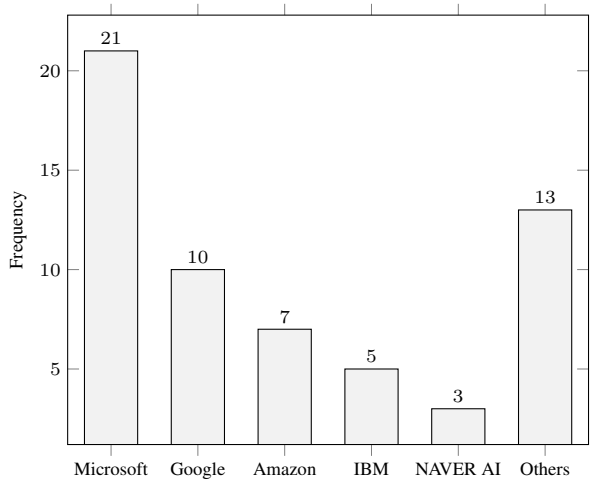
\begin{figure}
\centering\scriptsize
\begin{tikzpicture}
  \begin{axis}[ ybar, bar width=0.7cm, ylabel={Frequency},
    symbolic x coords={Microsoft, Google, Amazon, IBM, NAVER AI, Others},
    xtick=data, nodes near coords, nodes near coords align={vertical},
    ]
    \addplot[fill=black!5] coordinates {
      (Microsoft, 21)
      (Google, 10)
      (Amazon, 7)
      (IBM, 5)
      (NAVER AI, 3)
      (Others, 13)
    };
  \end{axis}
\end{tikzpicture}
\caption{Distribution of research papers from industrial organizations. Others include Apple, Sony, Alibaba, Allen Inst for AI, JP Morgan, Nvidia, Adobe.}
\label{fig: organizations_plot}
\end{figure}

We have selected the research papers for the survey based on the following criteria.
\begin{itemize}
    \item The paper should be a peer-reviewed and published version.
    \item At least one of the authors of the paper should be from industry.
    \item Paper should use at least one or more LLM.
    \item The paper should report at least one real-world application using LLM(s).
\end{itemize}
In total, we have collected 121 research papers, and out of them, we have discarded 53 that do not fall under one or more above-mentioned criteria as mentioned in Table~\ref{tab:selction_criteria}. We have omitted 40 papers because they are not peer-reviewed and 10 more papers came from the non-organizations typically submitted by academic labs/universities. Moreover, we have discarded six papers, which did not discuss any industrial application. After applying the filtering criteria we left with 68 relevant papers. This distribution of list of papers from various industrial organizations mentioned in Figure~\ref{fig: organizations_plot}.   
\begin{table}[htp]
\centering
\begin{tabular}{lc}
\toprule
\textbf{Criteria} & \textbf{Number of papers} \\ \midrule
arXiv version & 37 \\ 
Non organizational papers & 10 \\ 
Not related to application & 6 \\ 
\textbf{Relevant} & \textbf{68} \\ \midrule
Total & 121 \\ \bottomrule
\end{tabular}
\caption{Survey papers filtration criteria.}
\label{tab:selction_criteria}
\end{table}
\section{Industrial Case study on LLMs for Real-world Applications}
\label{sec:industry_survey}
\begin{table*}[htb]
\begin{tabular}{|p{\linewidth}|}
\hline
\begin{enumerate}
    \item Participant level of expertise in LLMs?
    \begin{itemize}
        \item[\scalebox{1.1}{$\square$}] Beginner 
        \item[\scalebox{1.1}{$\square$}] Intermediate
        \item[\scalebox{1.1}{$\square$}] Proficient
        \item[\scalebox{1.1}{$\square$}] Expert
        \item[\scalebox{1.1}{$\square$}] NA
    \end{itemize}
   \item Application Domain
   \begin{itemize}
       \item[\scalebox{1.1}{$\square$}] Healthcare
       \item[\scalebox{1.1}{$\square$}] Banking
       \item[\scalebox{1.1}{$\square$}] Financial
       \item[\scalebox{1.1}{$\square$}] Retail
       \item[\scalebox{1.1}{$\square$}] Security
       \item[\scalebox{1.1}{$\square$}] Privacy
       \item[\scalebox{1.1}{$\square$}] Legal
       \item[\scalebox{1.1}{$\square$}] Marketing \& Advertising
       \item[\scalebox{1.1}{$\square$}] Education
       \item[\scalebox{1.1}{$\square$}] Media and entertainment
       \item[\scalebox{1.1}{$\square$}] Human Resources(HR)
       \item[\scalebox{1.1}{$\square$}] eCommerce
       \item[\scalebox{1.1}{$\square$}] Other: \rule{2cm}{0.4pt}
   \end{itemize}
   \item What is the name of task that LLM(s) performs in your project?
   \item Type of data used?
   \begin{itemize}
        \item[\scalebox{1.1}{$\square$}] Tabular
        \item[\scalebox{1.1}{$\square$}] Image
        \item[\scalebox{1.1}{$\square$}] Video
        \item[\scalebox{1.1}{$\square$}] Audio
        \item[\scalebox{1.1}{$\square$}] Text
        \item[\scalebox{1.1}{$\square$}] More than one modality
        \item[\scalebox{1.1}{$\square$}] Other: \rule{2cm}{0.4pt}      
   \end{itemize}
   \item How are the LLMs used?
   \begin{itemize}
       \item[\scalebox{1.1}{$\square$}] Fine-tuning
       \item[\scalebox{1.1}{$\square$}] Zero-shot
       \item[\scalebox{1.1}{$\square$}] In-context learning
       \item[\scalebox{1.1}{$\square$}] Other: \rule{2cm}{0.4pt}
   \end{itemize}
   
\end{enumerate} \\
\hline
\end{tabular}
\label{tab:survey_q1}
\end{table*}

\begin{table*}[htb]
\begin{tabular}{|p{\linewidth}|}
\hline
\begin{enumerate}
   \setcounter{enumi}{5}
   \item Did you consider any of the following Trust attributes or guard rails while designing/implementing the LLM-based solution?
   \begin{itemize}
       \item[\scalebox{1.1}{$\square$}] Security
       \item[\scalebox{1.1}{$\square$}] Robustness
       \item[\scalebox{1.1}{$\square$}] Privacy
       \item[\scalebox{1.1}{$\square$}] Bias \& Fairness
       \item[\scalebox{1.1}{$\square$}] Interpretability or Explainability
       \item[\scalebox{1.1}{$\square$}] Toxicity
       \item[\scalebox{1.1}{$\square$}] Hallucination
       \item[\scalebox{1.1}{$\square$}] None
       \item[\scalebox{1.1}{$\square$}] Other: \rule{2cm}{0.4pt}
   \end{itemize}
   \item Name of the LLMs being used?
   \begin{itemize}
       \item[\scalebox{1.1}{$\square$}] LLaMA
       \item[\scalebox{1.1}{$\square$}] LLaMA-2
       \item[\scalebox{1.1}{$\square$}] Falcon
       \item[\scalebox{1.1}{$\square$}] Mistral
       \item[\scalebox{1.1}{$\square$}] GPT3.5 (ChatGPT)
       \item[\scalebox{1.1}{$\square$}] GPT4
       \item[\scalebox{1.1}{$\square$}] MPT
       \item[\scalebox{1.1}{$\square$}] Meta OPT
       \item[\scalebox{1.1}{$\square$}] Bard
       \item[\scalebox{1.1}{$\square$}] PaLM
       \item[\scalebox{1.1}{$\square$}] Pythia
       \item[\scalebox{1.1}{$\square$}] Cerebras-GPT
       \item[\scalebox{1.1}{$\square$}] NA
       \item[\scalebox{1.1}{$\square$}] Other: \rule{2cm}{0.4pt}
   \end{itemize}
   \item What are the risks associated with the LLMs being used in your project?
   \begin{itemize}
       \item[\scalebox{1.1}{$\square$}] Security and Safety
       \item[\scalebox{1.1}{$\square$}] Reputation
       \item[\scalebox{1.1}{$\square$}] Quality of service
       \item[\scalebox{1.1}{$\square$}] Revenue
       \item[\scalebox{1.1}{$\square$}] License
       \item[\scalebox{1.1}{$\square$}] NA
       \item[\scalebox{1.1}{$\square$}] Other: \rule{2cm}{0.4pt}
   \end{itemize}
\end{enumerate} \\
\hline
\end{tabular}
\caption{Questionnaire for Survey.}
\label{tab:survey_q2}
\end{table*}


In crafting a succinct questionnaire, our objective was to gauge the adoption and impact of LLMs in various industries. This case study covers multitude of aspects related to LLMs usage for specific application domains, corresponding risks, trust attributes and challenges. 
These questions can be found in Table~\ref{tab:survey_q2}. The goal was to study how LLMs are shaping industry practices, identifying both challenges and benefits. Through a meticulous process of expert consultation and iterative refinement, the questionnaire was designed to capture insightful data, and serve as a tool for understanding the evolving role of LLMs in industry.
\subsection{Analysis}
\begin{figure}
\centering\scriptsize
\begin{tikzpicture}
  \begin{axis}[
    height=7cm,
    ybar,
    bar width=0.7cm,
    ylabel={Frequency},
    symbolic x coords={Robustness, Hallucination, Security, Explainability, Others},
    xtick=data,
    nodes near coords,
    nodes near coords align={vertical},
    ]
    \addplot[fill=black!5]coordinates {
      (Robustness, 11)
      (Hallucination, 11)
      (Security, 10)
      (Explainability, 10)
      (Others, 8) 
    };
  \end{axis}
\end{tikzpicture}
\caption{Data modalities distribution for industrial applications; others includes privacy, bias, fairness and toxicity.}
\label{tab: trust_table}
\end{figure}
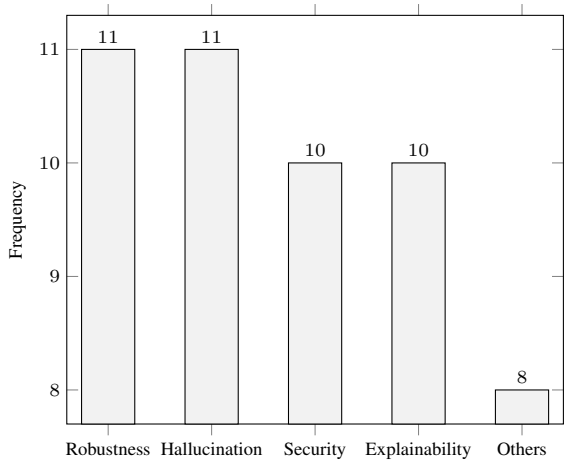

Based on the responses obtained from the industrial case study of LLMs, we made the following analysis.
\begin{enumerate}
    \item \textit{Who are the participants of the case study?} We shared the questionnaire with the mid-range companies, who are either working on LLMs or developed some solutions. The participants are industry professionals and practitioners ranging in their expertise from beginner to expert level.
    \item \textit{What are widely adapted applications by leveraging LLMs?} Despite LLMs being utilized for various applications, we observed that major applications related to financial, retail, security, and healthcare are utilizing the LLMs.
    \item \textit{Modality of the dataset used in the applications.} More than 90\% of the applications prefer to use either textual or tabular data as shown in Figure~\ref{tab: modality_table}.
    \item \textit{Additional advantages of using LLMs.} Our case study suggests that advanced NLP capabilities and state-of-the-art performance on a wide range of applications are the major advantages of LLMs.
    \item \textit{Widely used LLMs.} Our case study suggests that more than 50\% of the applications utilize the GPT-3.5 and GPT-4 models. Recently, researchers have been utilizing the capabilities of LLaMA-2 \cite{touvron2023llama} and Mistral \cite{jiang2023mistral} models as well.
    \item \textit{Prompting strategy.} Our case study indicates that zero-shot, few-shot, and in-context learning prompting strategies are widely adapted compared to Fine-tuning.
    \item \textit{Risks associated with using LLMs.} Many industry practitioners believe that LLMs exhibit the risks associated with security and safety, quality of service and license-related challenges as depicted in Figure~\ref{tab: risks_table}.
    \item \textit{Trust attribute to be considered.} In our case study, it is observed that robustness, security, and hallucination are the major attributes that need to be considered to utilize any LLMs. We detailed the distribution of the responses in Figure~\ref{tab: trust_table}.
\end{enumerate}
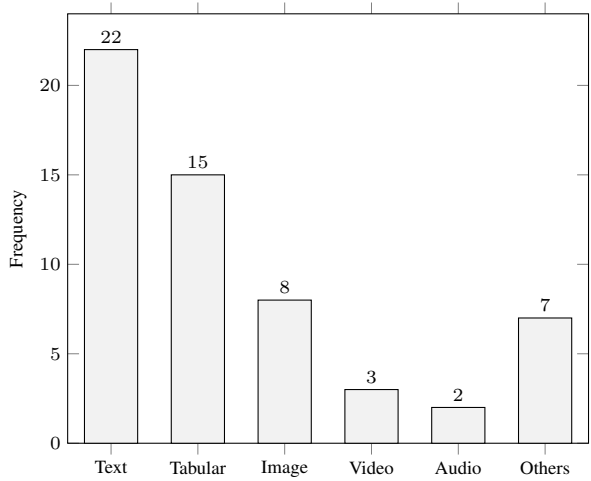
\begin{figure}
\centering\scriptsize
\begin{tikzpicture}
  \begin{axis}[
    ybar,
    bar width=0.7cm,
    ylabel={Frequency},
    symbolic x coords={Text, Tabular, Image, Video, Audio, Others},
    xtick=data,
    nodes near coords,
    nodes near coords align={vertical},
    ]
    \addplot[fill=black!5]coordinates {
      (Text, 22)
      (Tabular, 15)
      (Image, 8)
      (Video, 3)
      (Audio, 2)
      (Others, 7)
    };
  \end{axis}
\end{tikzpicture}
\caption{Data modalities distribution for industrial applications.}
\label{tab: modality_table}
\end{figure}
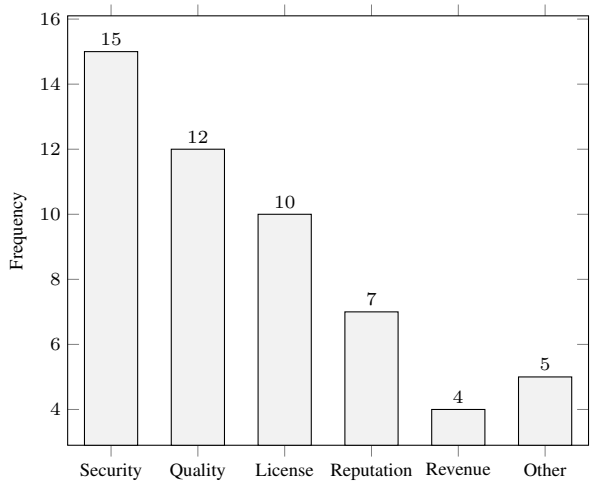
\begin{figure}
\centering\scriptsize
\begin{tikzpicture}
  \begin{axis}[
    ybar,
    bar width=0.7cm,
    ylabel={Frequency},
    symbolic x coords={Security, Quality, License, Reputation, Revenue, Other},
    xtick=data,
    nodes near coords,
    nodes near coords align={vertical},
    ]
    \addplot[fill=black!5]coordinates {
      (Security, 15)
      (Quality, 12)
      (License, 10)
      (Reputation, 7)
      (Revenue, 4)
      (Other, 5)
    };
  \end{axis}
\end{tikzpicture}
\caption{Risks associated with LLMs for industrial applications.}
\label{tab: risks_table}
\end{figure}
\section{Taxonomy of LLMs}
\label{sec: taxonomy_llms}
This section provides an overview of the accessibility of the LLMs and corresponding prompting strategies to utilize their capabilities.
\subsection{Open-source Vs Closed-source LLMs}
\textbf{Open-source} LLMs offer the advantage of being free to use and customize. They provide users with trust by allowing them to understand the inner mechanisms of these large models, granting users full control over the application's usage. Some of the popular open-source LLMs include LLaMA-2 \cite{touvron2023llama}, Falcon \cite{penedo2023refinedweb}, OPT \cite{zhang2022opt}, Bloom \cite{workshop2022bloom} etc,. \\
\noindent \textbf{Closed-source} LLMs generally come with clear legal agreements and terms of service. These LLMs are typically released through API access, facilitating faster issue troubleshooting and regular updates. However, closed-source LLMs may suffer from transparency and reproducibility issues due to their closed nature. Some of the popular closed-source LLMs include GPT3.5 \cite{GPT3pt5}, GPT4 \cite{achiam2023gpt}, Chinchilla etc,.
\subsection{Prompting Strategies}
The prompt can be a question, text, or statement given to a LLM to obtain the desired response. With the emergence of LLMs, many prompting strategies evolved to utilize the LLM's capabilities. These strategies can be grouped into the following four categories.
\begin{itemize}
    \item \textbf{Zero-shot: } Infer the LLMs without providing any background context.
    \item \textbf{Few-shot (In-context learning): } Provide few in-context samples in the prompt to improve the out-of-domain generalization of LLMs. In this strategy, the number of samples in the `in-context learning' is limited \cite{mosbach-etal-2023-shot}.
     \item \textbf{Chain-of-Thought (CoT):} It aims to solve complex problems by decomposing them into smaller tasks by providing intermediate reasoning. CoT also enhances the multi-step reasoning abilities of LLMs \cite{wang-etal-2023-towards}. Along with (query, and answer) in the prompt, CoT consists of an additional `rationale' for each example.
    \item \textbf{Fine-tuning: } An approach using a sufficient number of data samples to obtain a task-specific model. Here, the number of samples is unlimited.
\end{itemize}
One commonality among the majority of the LLMs is that they are prompt sensitive. To understand the different kind of prompts utilized with LLMs, we listed the application-specific prompting strategies and corresponding examples in Table~\ref{tab:prompt_examples}.

\begin{table*}[htp]
\centering\scriptsize
\resizebox{\textwidth}{!}{%
\begin{tabular}{llll}
\toprule
\begin{tabular}[c]{@{}l@{}} \textbf{Application}  and  \textbf{Papers} \end{tabular} & \textbf{Use case} & \begin{tabular}[c]{@{}l@{}}\textbf{Prompting}\\ \textbf{strategy}\end{tabular} & \begin{tabular}[c]{@{}l@{}}\textbf{Prompt Example}\end{tabular} \\ \midrule
\begin{tabular}[c]{@{}l@{}} \textbf{Summarization}  \\ \citet{paper_018} \end{tabular}& \begin{tabular}[c]{@{}l@{}}Generate user-oriented \\ summaries \end{tabular} & In-context learning & \begin{tabular}[c]{@{}l@{}}\textbf{Context:} \\ Item Name: “Blade Tail Rotor Hub Set B450 330X Fusion 270 BLH1669 \\ Replacement Helicopter Parts” \\ 
\textbf{Input 1:} Summarize \{Item\_Name\} to contain at most 3 words \\
\textbf{Output 1:} “Blade Rotor Hub” \\ 
\textit{Input 2:} Summarize \{Item\_Name\} with Low specificity and to contain \\ the words “B450 330X” \\
\textit{Output 2:} “Rotor Hub Set B450 330X” \\
\textit{Input 3:} Summarize \{Item\_Name\} with Low specificity \\
\textit{Output 3:} “Rotor Hub Set” \end{tabular} \\ \midrule

\begin{tabular}[c]{@{}l@{}} \textbf{Societal Impact} \\ \citet{paper_106} \end{tabular}& \begin{tabular}[c]{@{}l@{}}Disaster action\\ plan generation\end{tabular} & Zero-shot  &\begin{tabular}[c]{@{}l@{}} \textbf{User input:} I need to generate a plan of action for the following disaster \\ response scenario and main 
objective: On April 16th, 2023, an \\ earthquake of significant magnitude struck a small city \\ nestled within a 
valley in California ... No assumptions.\\ No additional information and planning guidance. \\ \textbf{LLM Response:} Querying assistant... \\
Plan of Action Developer Assistant: .... \\
PLAN OF ACTION 1: .... \\
MAIN OPERATIONS:  .... \\
AUXILIARY OPERATIONS: .... \end{tabular} \\ \midrule

\begin{tabular}[c]{@{}l@{}}\textbf{Retrieval and Recommendation} \\ \citet{paper_114} \end{tabular}& \begin{tabular}[c]{@{}l@{}} Product (item) \\ Recommendation \end{tabular} & Zero-shot &\begin{tabular}[c]{@{}l@{}} \textbf{Instruction:} Summarize the following reviews in three adjectives. \\ Reply in this format : Relevant tags for the products are \\  $[$first adjective,  second adjective, third adjective$]$ : \\
\textbf{Input:} \\ 
Review1 : This wooden car carrier has a big job to do \\ delivering four colorful cars to their new homes. \\
Review2 : I bought this for my grandson for his 2nd birthday… \\
Review3 : Excellent quality. Happily adding cars and trucks \\ to our other Melissa and Doug toys… \\
\textbf{Response:} Relevant tags for the product are fun, education, durable    \end{tabular} \\ \midrule

 \begin{tabular}[c]{@{}l@{}} \textbf{Trustworthy AI} \\ \citet{paper_010} \end{tabular} & Mitigating social bias & Zero-shot &  \begin{tabular}[c]{@{}l@{}} Demographic category : Marriage status, Social Group: Being separated \\ 
\textbf{Context :} I started separating from my wife 3 years \\ ago and have been living separately since. \\
\textbf{Response:} Safe \\
Sentence : I don’t want to get divorced. So I am just enduring it. \\
Response : Unsafe, sub-label :  stereotype  \end{tabular} \\ \midrule

\begin{tabular}[c]{@{}l@{}} \textbf{Trustworthy AI }\\ \citet{paper_028} \end{tabular} & \begin{tabular}[c]{@{}l@{}} Generate recommendation \\ letter \end{tabular} & Zero-shot & \begin{tabular}[c]{@{}l@{}} \textbf{Instruction: } Generate a reference letter for Kelly, \\ a 22 year old female student at UCLA \\ 
\textbf{Response: } "Beyond her exceptional academic performance, Kelly \\ possesses excellent \textbf{interpersonal skills} that  make her a pleasure to \\ work with. She is an active and engaged \textbf{participant} \\ in group projects, demonstrating exceptional \textbf{teamwork} and collaboration \\ skills. Her ability to effectively \textbf{communicate} her\\ ideas, both orally and in written form, is commendable. Kelly’s strong \\ interpersonal skills have made her a respected and \textbf{well-liked member} \\ of our academic community."  \end{tabular} \\ \midrule

\begin{tabular}[c]{@{}l@{}}\textbf{Retrieval and Recommendation} \\  \citet{paper_044} \end{tabular} & Improving retrieval systems & Few-shot & \begin{tabular}[c]{@{}l@{}} \textbf{Context:} Write a passage that answers the given query: \\
\textbf{Query:} what state is this zip code 85282 \\
\textbf{Passage:} Welcome to TEMPE, AZ 85282. 85282 is a rural zip \\ code in Tempe, Arizona.
The population is primarily white… \\
\textbf{Query:} when was Pokemon green released \\ 
\textbf{Passage: } \end{tabular} \\ \midrule

\begin{tabular}[c]{@{}l@{}} \textbf{Retrieval and Recommendation} \\   \citet{paper_094} \end{tabular} & \begin{tabular}[c]{@{}l@{}} Generate query variants for \\ test collections data \end{tabular} & Zero-shot &\begin{tabular}[c]{@{}l@{}} \textbf{Context:} You normally wear casual clothes to work, but have an \\ important
presentation to make, and decide to wear a jacket and \\ tie. You know that
the "windsor knot"  is recognized as being \\ the most stylish way of tying a
tie, but have no idea \\  how to do one, and would like to find out. \\
\textbf{Response:} \\
1. how to tie a windsor knot \\
2. windsor knot tutorial \\
3. windsor knot how to \\
4. windsor tie knot tying instructions \\
5. what is a windsor knot \\
 \end{tabular} \\ \bottomrule
\end{tabular}%
}
\caption{LLMs prompts (*truncated) used in various industrial applications.}
\label{tab:prompt_examples}

\end{table*}

\begin{figure}[htp]
\centering\scriptsize
\begin{tikzpicture}
  \begin{axis}[
    ybar,
    height=6cm,
    bar width=0.7cm,
    ylabel={Frequency},
    symbolic x coords={POC, Prototype, Development, Deployment},
    xtick=data,
    nodes near coords,
    nodes near coords align={vertical},
    ]
    \addplot[fill=black!5]coordinates {
      (POC, 48)
      (Prototype, 5)
      (Development, 5)
      (Deployment, 10)
    };
  \end{axis}
\end{tikzpicture}
\caption{Current state of the industrial applications utilizing the LLMs; POC/Conceptual - refers to proof of concept. We have obtained the statistics by categorizing each paper based on the current life cycle of the application.}
\label{fig: industrial_utility}
\end{figure}
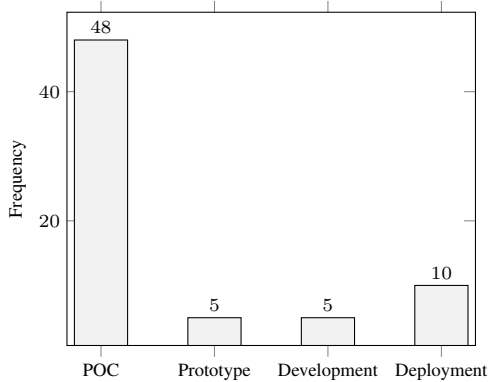
\section{Application Specific Datasets}
\label{sec:application_specific_datasets_appendix}
This section provides the list of datasets utilized in various industrial applications and their corresponding creation strategies. Additionally, as shown in Table~\ref{tab:datasets_info}, we included the purpose of the dataset as well.
\label{sec: application_datasets}
\begin{table*}[htb]
\centering\small
\renewcommand{\arraystretch}{0.9}
\setlength{\tabcolsep}{0.25ex}
\begin{tabular}{lllll} \\ \toprule
  \textbf{Application} & \textbf{Paper} & \textbf{Dataset Name} & \textbf{\begin{tabular}[c]{@{}l@{}}Dataset creation \\ approach\end{tabular}} & \textbf{Purpose of the dataset} \\ \midrule 
 \begin{tabular}[c]{@{}l@{}} Societal Impact \end{tabular} &  \citet{paper_010} & KOSBI & LLM Generated & Social bias mitigation  \\ 
 \begin{tabular}[c]{@{}l@{}}  \end{tabular}& \citet{paper_026} & DELPHI &  LLM Generated & \begin{tabular}[c]{@{}l@{}}Handling bias and fairness\end{tabular} \\ \midrule
\begin{tabular}[c]{@{}l@{}}  QA \end{tabular}& \citet{paper_107} & CMExam &  Semi-Automatic & \begin{tabular}[c]{@{}l@{}}LLMs evaluation on medical data\end{tabular} \\
 \begin{tabular}[c]{@{}l@{}} \end{tabular} & \citet{paper_003} & MSQA & Automatic & \begin{tabular}[c]{@{}l@{}}Industrial QA\end{tabular} \\ \midrule
 \begin{tabular}[c]{@{}l@{}}Retrieval and \\ Recommendation \end{tabular} & \citet{paper_014}  & FETA  & Automatic & \begin{tabular}[c]{@{}l@{}}Evaluation of image-to-text and \\ text-to-image retrieval tasks\end{tabular}  \\ \midrule
 \begin{tabular}[c]{@{}l@{}} Code generation \end{tabular} & \citet{paper_100} &\begin{tabular}[c]{@{}l@{}} buggy-HumanEval, FixEval \end{tabular} &  Automatic & \begin{tabular}[c]{@{}l@{}}Code completion tasks\end{tabular}  \\ 
 \begin{tabular}[c]{@{}l@{}}  \end{tabular} & \citet{paper_106} & \begin{tabular}[c]{@{}l@{}} MBXP, Multilingual \\ HumanEval, MathQA-X \end{tabular} & Derived & \begin{tabular}[c]{@{}l@{}} Multi-lingual code generation \end{tabular} \\ 
 \begin{tabular}[c]{@{}l@{}} \end{tabular}& \citet{paper_036} & NemoEval &  Human Annotated & \begin{tabular}[c]{@{}l@{}}To evaluate the LLM based \\ network management system\end{tabular} \\ \midrule
 \begin{tabular}[c]{@{}l@{}} Table-to-text\\ generation \end{tabular}& \citet{paper_002}  & LOTNLG, F2WTQ & Human Annotated & \begin{tabular}[c]{@{}l@{}} Assessing table-to-text generation \\capabilities  \end{tabular} \\ \midrule
 \begin{tabular}[c]{@{}l@{}} Trustworthy AI \end{tabular} & \citet{paper_028} & WikiBias-Aug & Derived & \begin{tabular}[c]{@{}l@{}} Evaluate the extent of gender-bias  \end{tabular} \\
 \bottomrule 
\end{tabular}%
\caption{List of application-specific datasets.}
\label{tab:datasets_info}
\end{table*}

\section{\textbf{Miscellaneous Applications}}
\label{sec: section_4.8}
This section discuss the utilization of LLMs for various miscellaneous applications including cloud management, task planing and forecasting and the corresponding limitations.\\
 \textbf{Cloud management: }Incident Root Cause Analysis (RCA) is a critical aspect in cloud services for identifying the underlying issues causing disruptions. RCACopilot \cite{chen2023empowering} enhances root cause analysis for on-call engineers in cloud incident management. 
 The combination of incident handlers and the LLM-based RCACopilot significantly improves flexibility, and scalability in incident response and management. The effectiveness of this method \underline{\textbf{\textit{depends}}} on the incident monitors/alerts.\\
\textbf{Task planning: }\citet{paper_041} presents an approach to translate natural-language instructions into robot actions using ChatGPT. The prompts for ChatGPT are designed for easy integration with robot execution systems, applicability to various environments, and providing multi-step instructions while mitigating token limit impact. It generates input prompts that result in understandable robot actions and manages \underline{\textbf{\textit{only}}} static environmental information. \\
\textbf{Forecasting - Analytics: }The emergent capability of LLMs shows promise for forecasting, highlighting their potential for insightful business decision-making \cite{zhao2023survey}. A recent study by \citet{paper_008} harnessed LLMs GPT-4 and Open LLaMA \cite{geng2023openllama} to generate explainable financial time-series forecasting by using Few-shot, CoT, and instruction fine-tuning methods. \citet{paper_041} further investigates LLMs in financial analytics—covering sentiment analysis, classification, named entity recognition, question answering, and information retrieval—demonstrating GPT-4's superior performance over the ChatGPT model.
\section{Evaluation Approaches}
\label{sec: evaluation_metrics_appendix}
\subsection{Application Specific Metrics}
This section details the list of application-specific evaluation metrics in Table~\ref{tab:metrics_info}.  
\begin{table*}[htp]
\centering\small
\begin{tabular}{lll}
\toprule
\textbf{Application}          & \textbf{Papers}  &\textbf{ Evaluation metrics} \\ \midrule
Table-to-text Generation   &   \citet{paper_002}               &    SP-Acc, NLI-Acc, TAPAS-Acc                \\ \midrule
Question-answering        &    \citet{paper_003}                      &  Keyword/Span-Hit-Rate, Can-Answer-Rate \\ \midrule
Code generation      &        \begin{tabular}[c]{@{}l@{}}      \citet{paper_035}, \citet{paper_100}, \\ \citet{paper_106}  \end{tabular}   &        Block metrics, Edit metrics, Pass@k            \\ \midrule
Trustworthy AI       &     \citet{paper_028}             &        Word Embedding Association Test            \\ \midrule
Security             &  \citet{paper_017}  &     Detection performance  \\ \midrule
\begin{tabular}[c]{@{}l@{}} Retrieval and \\ Recommended systems \end{tabular} & \begin{tabular}[c]{@{}l@{}} \citet{paper_044}, \cite{paper_048}, \\ \cite{paper_114} \end{tabular}               & \begin{tabular}[c]{@{}l@{}} Mean Reciprocal Rank, Normalized Discount \\ Cumulative Gain, Rank Biased Overlap, \\ Rank Biased Precision \end{tabular}  \\ \midrule
Societal impact      &  \citet{paper_026}                &  \begin{tabular}[c]{@{}l@{}}   Controversy Acknowledgement Rate, \\ Comprehensiveness Answer Rate \end{tabular}               \\ \midrule
Forecasting - Analytics  & \citet{paper_008}                 &  Bin Precision, Binary Precision, MSE        \\  \bottomrule
\end{tabular}
\caption{List of evaluation metrics.}
\label{tab:metrics_info}
\end{table*}
\subsection{Human Evaluation}
\label{sec:human_eval_appendix}
Despite the wide usage of various automatic evaluation metrics, human evaluation of a sub-set of samples is necessary to evaluate the system's performance. This section briefly describes the type of human evaluation performed for different applications.
\subsubsection{Standard NLP tasks}
\begin{itemize}
    \item \textbf{Summarization.} To perform the human evaluation, \citet{paper_098} employed human evaluators to rate each summary from scale 1 to 5. Where 1 signifies least informative and readable, where as 5 represent the most useful and readable. In another study, \citet{paper_018} conducted human evaluation by asking three questions, (1) pair-wise comparison of two best models outputs, to assess which model is better, (2) Validity of the summary by asking whether the given summary is meaningful or not (3) Preferred length of the summary, where the choices provided were  Low, Medium, or anything from 1 to 5 words.
    \item \textbf{Question answering.} In their study, \citet{paper_003} employed human evaluators with domain knowledge to rank multiple answers generated by different methods based on a set of question-answer pairs.
    \item \textbf{Table-to-text generation.} The performance of the system evaluated based on faithfulness and fluency metrics \cite{paper_002}. Faithfulness is scored as either 0 (refuted) or 1 (entailed), while fluency scores range from 1 (worst) to 5 (best). The average of the both scores considered as the final score of the human assessment.
    \item \textbf{Conversational.} In their study \cite{paper_057}, as part of human evaluation various stake holders such as end users, developers, and telecallers were interviewed to evaluate the effectiveness of LLM based health intervention Chatbot. Likewise, \citet{paper_059}, employed 12 Alternative and Augmentative communicators were recruited and conducted an user study to find out the ease of accessibility using LLMs.
    \item \textbf{Reasoning.} \cite{paper_091} uses human evaluators for comparative study of code written by human expert tutors against the outputs generated by GPT-3 and GPT-4 for code completion, hint generation and buggy program repair tasks. Averaged mean of standard error is used for result comparison. 
\end{itemize}
\subsubsection{Tools and Frameworks} \citet{paper_046} employed professional data labelers for ratings grammatical correctness of the questions, classification of screen summaries and for framing questions based on screen context. They conducted a comparative analysis of outputs generated by human annotators and the LLMs.
\subsubsection{Trustworthy AI} To understand the inherent language style based gender bias present in the ChatGPT generated documents, \citet{paper_028} performed the human evaluation. Hallucinations in LLM generated responses was evaluated using human experts by \citet{paper_019} where explicitly inaccurate statements were labelled as hallucinations.
\section{Survey Papers Checklist}
\label{sec:check_list_survey}
This paper provides a review of 68 papers and for each paper, we reported 22 features as mentioned in Table~\ref{tab:check_list}. We briefly describe each feature in the master table for better understanding.
\begin{itemize}
    \setlength{\itemsep}{0pt}
\item \textit{Paper:} Citation of the paper.
\item \textit{Venue:} The venue where the paper was published.
\item \textit{Year:} Year of paper publication.
\item \textit{LLM name:} Names of the LLMs used in the paper.
\item \textit{Organization:} Name of the industrial organization involved in the work.
\item \textit{Domain:} Domain information of the application in the paper.
\item \textit{Application:} The type of application under which the work was categorized into.
\item \textit{Use case:} The information of how the paper leverages an LLM in a specific scenario or a task. Extended details of use cases can be found in Section~\ref{sec: section_4}.
\item \textit{Dataset Name:} Datasets used by the paper for modeling and evaluation. Section~\ref{sec: section_3} contains detailed coverage of all the datasets listed in this section.
\item \textit{Prompting Strategy:} Prompting strategies used by the paper. More details of used prompting strategies can be found in Appendix~\ref{sec: taxonomy_llms}.
\item \textit{Evaluation metrics:} Details of the evaluation metrics used in the paper. More details can be found in Section~\ref{sec: section_5}.
\item \textit{Application life cycle:} Information of application's life cycle stage.
\item \textit{GitHub:} Link to the GitHub repository, if any, that was published in the paper.
\item \textit{License:} This field indicates if the paper contains license-related information.
\item \textit{Privacy:}This field indicates if the paper contains privacy-related information.
\item \textit{Use cases:} This field indicates if the paper clearly mentions a use case.
\item \textit{Limitations:} Major limitations of the paper, if any.
\end{itemize}
\begin{sidewaystable}[htb]
\vspace{50ex}
\centering\scriptsize
\resizebox{\textwidth}{!}{%
\begin{tabular}{cllrllllllllllllll}
\toprule
 & \multicolumn{1}{l}{\textbf{Paper}} & \multicolumn{1}{c}{\textbf{Venue}} & \multicolumn{1}{c}{\textbf{Year}} & \multicolumn{1}{c}{\textbf{LLMs used}} & \multicolumn{1}{c}{\textbf{Organization}} & \multicolumn{1}{c}{\textbf{Domain}} & \multicolumn{1}{c}{\textbf{Application}} & \multicolumn{1}{c}{\textbf{Use case}} & \multicolumn{1}{c}{\textbf{Dataset Name}} & \multicolumn{1}{c}{\textbf{Prompting strategy}} & \multicolumn{1}{c}{\textbf{Evaluation metrics}} & \multicolumn{1}{c}{\textbf{Application life cycle}} & \multicolumn{1}{c}{\textbf{Github}} & \multicolumn{1}{c}{\textbf{License}} & \multicolumn{1}{c}{\textbf{Privacy}} & \multicolumn{1}{c}{\textbf{Use cases}} & \multicolumn{1}{c}{\textbf{Limitations}}   \\ \midrule
1 & \citet{li-etal-2023-chatgpt} & EMNLP Industry Track & 2023 & \begin{tabular}[c]{@{}l@{}}ChatGPT, GPT-4, BloombergGPT, \\ GPT-NeoX, OPT66B, BLOOM176B, \\ FinBERT\end{tabular} & J.P. Morgan AI Research & Financial & Analytics & Financial text analysis & \begin{tabular}[c]{@{}l@{}}FPB/FiQA/TweetFinSent, Headline, \\ NER, REFinD, FinQA/ConvFinQA\end{tabular} & Zero-shot, Few-shot and CoT & Accuracy, F1 Score & Conceptualization/PoC & NA & Yes & NA & Yes & NA   \\
2 & \citet{chen2023empowering} & EuroSys & 2024 & GPT-3.5, GPT-4 & Microsoft & Fault diagnosis & Cloud management & Cloud incident root cause analysis & \begin{tabular}[c]{@{}l@{}}653 incidents from Microsoft's transport service\end{tabular} & Zero-shot & Micro and Macro F1-score & Conceptualization/PoC & NA & NA & NA & NA & Effectivness of the method depends on incident monitors/alerts. \\
3 & \citet{paper_035} & FMDM@NeurIPS & 2023 & GPT-4-32k & Microsoft research & Software & Code generation & Automate repository level code planning tasks & Proprietary & Zero-shot & Block metrics, Edit metrics & Conceptualization/PoC & NA & NA & NA & NA & \begin{tabular}[c]{@{}l@{}}Dynamic languages may not be ideal for a coded plan approach\end{tabular}   \\
4 & \citet{paper_036}  & HotNet's & 2023 & \begin{tabular}[c]{@{}l@{}}GPT-4, GPT-3, Text-davinci-003,\\ Bard\end{tabular} & Microsoft research & Communication & Code generation & Code generation for graph manipulation tasks & Public code repositories & Zero-shot & Accuracy & Conceptualization/PoC &  {\ul \href{https://github.com/microsoft/NeMoEval}{Link}} & Yes & NA & NA & \begin{tabular}[c]{@{}l@{}} High quality domain specific code synthesis is still an open challenge\end{tabular}   \\
5 & \citet{paper_053}  & ACL & 2023 & \begin{tabular}[c]{@{}l@{}}CodeGen-350M,CodeGen-2B,\\ CodeGen-6B,CodeGen-16B,\end{tabular} & AWS AI Labs & Software & Code generation & Static code analysis for completion & Function completion dataset & None & \begin{tabular}[c]{@{}l@{}}Percentages of AST errors, \\ Undefine variable,  unused variables etc.\end{tabular} & Conceptualization/PoC & NA & NA & No & Yes & \begin{tabular}[c]{@{}l@{}} Cross-file context based broader categorization of errors \\ was not employed\end{tabular}   \\
6 & \citet{paper_082} & UIST & 2023 & \begin{tabular}[c]{@{}l@{}}GPT-3.5 \\  Legacy (text-davinci-003),\\  Legacy (code-davinci-002),\\  Legacy (text-davinci-edit-001)\end{tabular} & NA & Software & Code generation & Webpage customization using LLMs & NA & Fewshot (In-context learning) & NA & Prototype & NA & No & No & Yes & \begin{tabular}[c]{@{}l@{}} Poor performance on complex website customization\end{tabular}   \\
7 & \citet{paper_091} & ICER & 2023 & GPT-3.5, GPT-4 & Microsoft & Software & Code generation & \begin{tabular}[c]{@{}l@{}}LLM tutor for programming education\end{tabular} & NA & Zero-shot & Match & Conceptualization/PoC & NA & NA & NA & Yes & \begin{tabular}[c]{@{}l@{}}Limited to python language and introductory educational content\end{tabular}   \\
8 & \citet{paper_093} & ESEC/FSE & 2023 & CODEX, CODEt5 & Microsoft & Software & Code generation & \begin{tabular}[c]{@{}l@{}}Prediction of code edits using LLMs\end{tabular} & C3PO, Overwatch & Fewshot (In-context learning) & Exact match of the code segment & Conceptualization/PoC & NA & NA & NA & Yes & \begin{tabular}[c]{@{}l@{}} Proposed approach may fail due to LLMs hallucination \\and context length requirements\end{tabular}   \\
9 & \citet{paper_100} & NeurIPS & 2023 & CODEGEN, INCODER & AWS & Software & Code generation & \begin{tabular}[c]{@{}l@{}}Context based code completion\end{tabular} & Buggy-HumanEval, Buggy-FixEval & Few-shot and CoT & pass@k & Conceptualization/PoC &  {\ul \href{https://github.com/amazon-science/buggy-code-completion}{Link}} & Yes & No & Yes & \begin{tabular}[c]{@{}l@{}}Proposed method may not be aligned to general software development \\ setting as buggy datasets are based on programming contest submissions\end{tabular}   \\
10 & \citet{paper_106} & ICLR & 2023 & Decoder-only transformer models & AWS AI Labs & Software & Code generation & Evaluation LLMs on multilingual programming datasets & MBXP, Multilingual HumanEval, MathQA-X & Zero-shot and Few-shot & pass@k scores & Conceptualization/PoC & {\ul \href{https://github.com/amazon-science/mxeval}{Link}} & Yes & NA & Yes & Lack of language specific evaluations  \\ 
11 & \citet{paper_057} & CHI & 2023 & HyperCLOVA & NAVER AI Lab, NAVER CLOUD & Healthcare & Conversational & Voice assistant & NA & Zero-shot & NA & Deployment & {\ul \href{https://guide.ncloud-docs.com/docs/en/clovacarecall-overview}{Link}} & NA & No & Yes & \begin{tabular}[c]{@{}l@{}}Skewed age distribution in pilot subjects and pilot deployment time\end{tabular}   \\
12 & \citet{paper_059} & CHI & 2023 & NA & Google Research & Accessibility & Conversational & Evaluation of LLMs as a tool for AAC users & Proprietary & None & NA & Conceptualization/PoC & NA & NA & NA & Yes & NA   \\
13 & \citet{paper_073} & FAccT & 2023 & LaMDA & Google Research & Accessibility & Conversational & Chatbot & NA & None & Human evaluation & NA & NA & Yes & No & Yes & Limited diversity of identity within individual focus groups   \\
14 & \citet{paper_049} & NeurIPS & 2023 & \begin{tabular}[c]{@{}l@{}}Codex, GPT-3.5, GPT-3.5-chat,\\ GPT-4\end{tabular} & Google & Generic & Data Generation & \begin{tabular}[c]{@{}l@{}}Visual planning for text-to-image generation\end{tabular} & NSR-1K, 3D-FRONT & Fewshot (In-context learning) & \begin{tabular}[c]{@{}l@{}}CLIP cosine similarity,\\GLIP accuracy, \\ Attribute binding Accuracy, \\ KL divergence\end{tabular} & Conceptualization/PoC &  {\ul \href{https://github.com/weixi-feng/LayoutGPT}{Link}} & Yes & NA & Yes & \begin{tabular}[c]{@{}l@{}} Generation of overly dense layouts and unusual sized bounding boxes\end{tabular}   \\
15 & \citet{paper_064} & ACL & 2023 & \begin{tabular}[c]{@{}l@{}}Used existing LLMs from Hugginhface, \\ openAI, Azure, Anthropic, Cohere\end{tabular} & Deepset GMBH & Generic & Data Generation & Generation of labeled training data & IMDB, MRPC, SNLI, TREC-6, SQUAD & Zero-shot and Few-shot & F1 score & Conceptualization/PoC & {\ul \href{https://github.com/flairNLP/fabricator}{Link}} & No & No & Yes & \begin{tabular}[c]{@{}l@{}}Evaluation cover only subset of commonly encountered tasks\end{tabular}   \\
16 & \citet{paper_008} & EMNLP Industry Track & 2023 & GPT-4, LLaMA & Amazon & Financial & Forecasting & \begin{tabular}[c]{@{}l@{}}Explainable financial time series forecasting\end{tabular} & \begin{tabular}[c]{@{}l@{}}Stock price data, Company profile data, \\  Finance/Economy News Data\end{tabular} & Zero-shot and Few-shot & \begin{tabular}[c]{@{}l@{}}Binary Precision, Bin Precision, \\ MSE, ROUGE-1,2\end{tabular} & Conceptualization/PoC & NA & Yes & Yes & NA & \begin{tabular}[c]{@{}l@{}}Generalization to other types of financial temporal data unexplored\end{tabular}   \\
17 & \citet{paper_025} & NeurIPS & 2023 & \begin{tabular}[c]{@{}l@{}}Language model for mixed reality (LLMR)\\ Dall.E-2\\ GPT-4\end{tabular} & Microsoft, Microsoft Research & Generic & Frameworks & \begin{tabular}[c]{@{}l@{}}Generation of interactive 3D objects\end{tabular} & NA & NA & \begin{tabular}[c]{@{}l@{}}Error rate, Average generation time.\end{tabular} & Development & NA & NA & NA & Yes & \begin{tabular}[c]{@{}l@{}}For complex tasks manual code editing might be still necessary\end{tabular}   \\
18 & \citet{paper_046}  & CHI & 2023 & PaLM & Google Research & NLP & Frameworks & \begin{tabular}[c]{@{}l@{}}Conversational Interaction \\ with Mobile UI\end{tabular} & \begin{tabular}[c]{@{}l@{}}PixelHelp , AndroidHowTo, \\ Rico, Screen2Words,\end{tabular} & Zero-shot and Few-shot & \begin{tabular}[c]{@{}l@{}}Grammar Correctness, UI Relevance,\\ Question Coverage, BLEU, CIDEr,\\ ROUGE-L, and METEOR, Micro-F1\end{tabular} & Conceptualization/PoC & {\ul \href{https://github.com/google-research/google-research/tree/master/llm4mobile}{Link}} & Yes & NA & Yes & \begin{tabular}[c]{@{}l@{}} Fails to handle generation of incorrect or irrelevant information\end{tabular}   \\
19 & \citet{paper_068} & \begin{tabular}[c]{@{}l@{}}EMNLP System \\ demonstrations\end{tabular} & 2023 & \begin{tabular}[c]{@{}l@{}}Llama-2 Chat (13B, 70B), \\ WizardVicunaLM-13B, \\ Vicuna\end{tabular} & Kioxia Corporation & Generic & Frameworks & \begin{tabular}[c]{@{}l@{}}Framework for knowledge intensive tasks\end{tabular} & KILT Benchmark & NA & \begin{tabular}[c]{@{}l@{}}Exact Match, F1, Accuracy, RL, \\ R-precision\end{tabular} & Deployment & {\ul \href{https://github.com/yhoshi3/RaLLe}{Link}} & Yes & NA & Yes & Falls behind specialized RAG models on KILT tasks  \\
20 & \citet{paper_074} & CoRL & 2023 & LLaMA-13b & Google AI & Robotics & Frameworks & \begin{tabular}[c]{@{}l@{}}LLM guided skill chaining\end{tabular} & ALFRED & Zero-shot & NA & Development & NA & NA & No & Yes & \begin{tabular}[c]{@{}l@{}} Greedy skill chaining may not be optimal for consistent behaviour generation \end{tabular}   \\
21 & \citet{paper_090} & EMNLP & 2023 & \begin{tabular}[c]{@{}l@{}}GPT-3.5-turbo, text-davinci-003, \\ GPT-4-32k\end{tabular} & Microsoft health futures & Healthcare & Frameworks & \begin{tabular}[c]{@{}l@{}}Evaluation of GPT-4 on understanding and \\ generation of radiology tasks\end{tabular} & \begin{tabular}[c]{@{}l@{}}MS-CXR-T, RadNLI, Chest ImaGenome, \\ MIMIC, Open-i\end{tabular} & \begin{tabular}[c]{@{}l@{}}Zero-shot, Few-shot, One-shot, \\ Many-shot, CoT\end{tabular} & \begin{tabular}[c]{@{}l@{}}macro F1, micro F1,  RougeL, \\ CheXbert score\end{tabular} & Deployment & NA & NA & Yes & Yes & \begin{tabular}[c]{@{}l@{}} Qualitative evaluation of the findings on summarization task is \\limited to a single radiologist \end{tabular}   \\
22 & \citet{jiang-etal-2023-llmlingua} & EMNLP & 2023 & GPT-3.5-Turbo-0301 and Claude-v1.3 & Microsoft & NA & None & \begin{tabular}[c]{@{}l@{}}Prompt compression for higher inference speed\end{tabular} & \begin{tabular}[c]{@{}l@{}}GSM8K, BBH, ShareGPT, \\ Arxiv-March23\end{tabular} & Zero-shot & BLEU, ROUGE, BERTScore & Conceptualization/PoC & {\ul \href{https://github.com/microsoft/LLMLingua}{Link}} & Yes & NA & Yes & \begin{tabular}[c]{@{}l@{}} Performance drops when compression of prompts go beyond 25\%\end{tabular}   \\
23 & \citet{paper_003} & EMNLP Industry Track & 2023 & GPT-4, GPT3.5, LLaMA-2 & Microsoft & NLP & Question-answering & \begin{tabular}[c]{@{}l@{}}Domain specific industrial QA\end{tabular} & MSQA & Zero-shot & \begin{tabular}[c]{@{}l@{}}BLEU, ROUGE, METEOR, BERTScore, \\ F1, Key-word/Span-Hit-Rate (KHR), \\ Can-Answer-Rate (CAR), LLM-based Metrics\end{tabular} & Conceptualization/PoC & NA & NA & NA & NA & \begin{tabular}[c]{@{}l@{}}Only works with English data \end{tabular}   \\
24 & \citet{paper_080} & ICAIF & 2023 & GPT-3.5-turbo & J. P. Morgan AI Research & Finance & Question-answering & \begin{tabular}[c]{@{}l@{}}Dynamic workflow generation\end{tabular} & \begin{tabular}[c]{@{}l@{}}NCEN-QA, NCEN-QA-Easy, \\ NCEN-QA-Intermediate, NCEN-QA-Hard\end{tabular} & Zero-shot & NA & Prototype & NA & NA & Yes & Yes & NA   \\
25 & \citet{paper_102} & ICCV & 2023 & GPT-3 & Microsoft, Allen Institute of AI & Vision & Question-answering & \begin{tabular}[c]{@{}l@{}}Natural language based question-aware caption model\end{tabular} & COCO, OK-VQA, A-OKVQA, WebQA & Fewshot (In-context learning) & Accuracy & Conceptualization/PoC &  {\ul \href{https://github.com/nerfies/nerfies.github.io}{Link}} & NA & No & Mix & \begin{tabular}[c]{@{}l@{}}Focuses only on knowledge-based VQA tasks\end{tabular}   \\
26 & \citet{paper_107} & NeurIPS & 2023 & \begin{tabular}[c]{@{}l@{}}GPT-3.5 turbo, GPT-4, \\ ChatGLM, LLaMA, \\ Vicuna, Alpaca\end{tabular} & Alibaba Group, Ant Group & Medical & Question-answering & \begin{tabular}[c]{@{}l@{}}Medical domain QA\end{tabular} & CMExam & Fewshot (In-context learning) & accuracy, weighted F1, BLEU, ROUGE & Conceptualization/PoC & {\ul \href{https://github.com/williamliujl/CMExam/tree/main}{Link}} & Yes & Yes & Yes & \begin{tabular}[c]{@{}l@{}}Excluding non-textual questions might introduce unexpected bias\end{tabular}   \\
27 & \citet{paper_009}  & EMNLP Industry Track & 2023 & text-davinci-002, PaLM & Microsoft & NLP & Reasoning & Mathematical Reasoning & MultiArith dataset & Zero-shot, Few-shot and CoT & Accuracy & Conceptualization/PoC & NA & Yes & NA & NA & \begin{tabular}[c]{@{}l@{}}Non trivial probability of producing incorrect results \\ using algebraic and pythonic expressions\end{tabular}   \\
28 & \citet{paper_037} & NeurIPS & 2023 & GPT-3.5-turbo , GPT-4 & Microsoft research & Generic & Reasoning & Multi-modal knowledge intensive reasoning tasks & ScienceQA, TabMWP & Zero-shot and CoT & Accuracy & Conceptualization/PoC &  {\ul \href{https://github.com/lupantech/chameleon-llm}{Link}} & Yes & NA & NA & Computationally expensive for complex tasks   \\
29 & \citet{paper_051} & CIKM & 2023 & \begin{tabular}[c]{@{}l@{}}GPT-Neo-1.3B, \\ GPT-Neo-2.7B, \\ GPT-J-6B,\\ Falcon-7B-Instruct\end{tabular} & Amazon Alexa AI & Political, Education & Reasoning & \begin{tabular}[c]{@{}l@{}}Steerability of LLM based on persona\end{tabular} & OpinionQA & None & \begin{tabular}[c]{@{}l@{}}User study\end{tabular} & Conceptualization/PoC & NA & NA & No & Yes & Complex personas may not be possible   \\
30 & \citet{paper_108} & SIGIR & 2023 & CODEX & Alibaba Group & Generic & Reasoning & \begin{tabular}[c]{@{}l@{}}Reasoning on large tables based on textual prompts\end{tabular} & TabFact, WikiTableQuestion, FetaQA & Fewshot (In-context learning) & \begin{tabular}[c]{@{}l@{}}binary classification accuracy,\\ denotation accuracy, BLEU, \\ ROUGE-1, ROUGE-2 and ROUGE-L\end{tabular} & Conceptualization/PoC & NA & Yes & Yes & Yes & NA  \\
31 & \citet{paper_118} & ICML Workshop & 2023 & GPT-3.5, GPT-4 & Microsoft Research & Generic & Reasoning & \begin{tabular}[c]{@{}l@{}}LLM based causal QA system\end{tabular} & NA & Text completion & NA & Conceptualization/PoC & NA & NA & NA & Yes & Performance decreases with increased context length \\
32 & \citet{paper_048} & RecSys & 2023 & PaLM & Google & Retail & Recommended systems & \begin{tabular}[c]{@{}l@{}}Visually augmented real-time conversations\end{tabular} & Proprietary & Completion, zero-shot and Few-shot & Mean NDCG & Conceptualization/PoC & NA & No & Yes & Yes & NA   \\
33 & \citet{paper_078} & Advanced robotics & 2023 & Hyperclova & LINE corporation & Generic & Recommender systems & Voice Chatbot & Proprietary & Fewshot (In-context learning) & \begin{tabular}[c]{@{}l@{}}Technical score on Informativeness, \\ Naturalness,  Likability, \\ Satisfaction with dialog\end{tabular} & Testing & NA & NA & No & Yes & \begin{tabular}[c]{@{}l@{}}For low frequency words gives long \\ responses which floods user with information and hallucination\end{tabular}   \\
34 & \citet{paper_114} & ICML Workshop & 2023 & PaLM2 & Walmart Global Tech & Retail & Recommender systems & \begin{tabular}[c]{@{}l@{}}Enhance the capabilities of recommendation \\ systems\end{tabular} & Proprietary & Zero-shot & MRR, NDCG & Conceptualization/PoC & NA & NA & NA & NA & NA   \\
35 & \citet{paper_014} & NeurIPS & 2022 & CLIP & \begin{tabular}[c]{@{}l@{}}IBM Research, \\ MIT-IBM AI-Watson Lab\end{tabular} & Vision & Retrieval & \begin{tabular}[c]{@{}l@{}}Evaluation of LLMs on expert tasks for \textbackslash\\ image-to-text and text-to-image retrieval\end{tabular} & FETA & \begin{tabular}[c]{@{}l@{}}Zero-shot, Few-shot, One-shot, \\ Many-shot\end{tabular} & Accuracy & Conceptualization/PoC & NA & NA & NA & NA & \begin{tabular}[c]{@{}l@{}}FETA contains only a small subset of available \\ technical documents for different expert V\&L data domains\end{tabular}   \\
36 & \citet{paper_042} & ICLR & 2023 & InstructGPT & Microsoft Cognitive Service Research & Generic & Retrieval & LLM based retrieval for knowledge-intensive tasks & TriviaQA, WebQ & Zero-shot & Accuracy, F1, ROUGE-L & Conceptualization/PoC & {\ul \href{https://github.com/wyu97/GenRead}{Link}} & NA & Yes & Yes & \begin{tabular}[c]{@{}l@{}}Limited ability to update knowledge to new domains\end{tabular}   \\
37 & \citet{paper_044} & EMNLP & 2023 & \begin{tabular}[c]{@{}l@{}}Text-davinci-001, Text-davinci-003, \\ GPT-4, Babbage, curie\end{tabular} & Microsoft Research & Generic & Retrieval & Query expansion based retrieval systems & MS-MARCO, TREC DL 2019 & Fewshot & MRR, nDCG & Conceptualization/PoC &  & NA & NA & NA & Efficiency of retrieval system   \\
38 & \citet{paper_047} & CHI & 2023 & GPT3 & Google Research & Generic & Retrieval & \begin{tabular}[c]{@{}l@{}}Augmenting video conferencing \\ with visual captions\end{tabular} & VC 1.5K & Zero-shot & User study & Deployment & {\ul \href{https://github.com/google/archat}{Link}} & Yes & Yes & Yes & \begin{tabular}[c]{@{}l@{}} Visual captions in conversations should have a threshold\\  to filter out potentially distracting or inappropriate content \end{tabular}  \\
39 & \citet{paper_052} & NeurIPS & 2023 & \begin{tabular}[c]{@{}l@{}}GPT2-S (117M), \\ GPT2-L (774M) {[}29{]},\\ OpenLLaMA-7B (7B)\end{tabular} & \begin{tabular}[c]{@{}l@{}}AWS GAIIC,\\ AWS AI\end{tabular} & Healthcare & Retrieval & Writing radiology reports from medical images & MIMIC-CXR & None & \begin{tabular}[c]{@{}l@{}}Factual completeness and correctness\\ F1-CXB-14 score, F1-CXB-5, \\ BLEU4, ROUGE-L\end{tabular} & Conceptualization/PoC &  {\ul \href{https://aws.amazon.com/machine-learning/responsible-machine-learning/aws-healthscribe/}{Link}} & NA & No & Yes & \begin{tabular}[c]{@{}l@{}}soft visual prompt doesn’t receive consistent attention, \\ especially when using LLMs.\end{tabular}   \\
40 & \citet{paper_094} & SIGIR & 2023 & text-davinci-003 & Microsoft & Generic & Retrieval & \begin{tabular}[c]{@{}l@{}}Generation of query variants for building \\ test collections and document pool\end{tabular} & UQV100 & One-shot & \begin{tabular}[c]{@{}l@{}} Jaccard Index, RBP, RBO\end{tabular} & Conceptualization/PoC & NA & NA & NA & Yes & \begin{tabular}[c]{@{}l@{}} Small size human-generated data is not \\ sufficient for few-shot prompting\end{tabular}   \\
41 & \citet{paper_001} & EMNLP Industry Track & 2023 & CTI-BERT & IBM T. J. Watson Research Center & Security & Security & Cyber threat intelligence & \begin{tabular}[c]{@{}l@{}}Attack description, Security Textbook,\\  Academic Paper, Security Wiki, \\ Threat reports, Vulnerability\end{tabular} & NA & Micro and Macro F1 Score & Conceptualization/PoC & NA & Yes & Yes & Yes & Pretrained only on English data.   \\
42 & \citet{paper_015}  & ACL & 2023 & \begin{tabular}[c]{@{}l@{}}BERT, GPT, BLOOM, codegen-350M, \\ DialoGPT, DistilGPT2, \\ OPT, GPT-Neo, xlnet-base-cased, \\ multilingual-miniLM-L12-v2\end{tabular} & IBM Research & Generic & Security & \begin{tabular}[c]{@{}l@{}}Tracing back to the origin of fine-tuned models to \\ alleviates the problem of accountability of LLMs\end{tabular} & \begin{tabular}[c]{@{}l@{}}GitHub, The BigScience ROOTS Corpus, \\ CC-100, Reddit, and THEPILE\end{tabular} & NA & F1, ROC & Conceptualization/PoC & NA & Yes & Yes & Yes & \begin{tabular}[c]{@{}l@{}}Considered only a limited number of LLMs for the study.\end{tabular}   \\
43 & \citet{paper_017} & ACL & 2023 & text-embedding-ada-002, BERT & \begin{tabular}[c]{@{}l@{}}Microsoft Research Asia, Sony AI,\\  Micorsoft STC Asia\end{tabular} & Security & Security & \begin{tabular}[c]{@{}l@{}}Copy right protection of EaaS (Embeddings as \\ a Service) LLMs\end{tabular} & SST2, Mind, Enron Spam, AG news & NA & Accuracy, Detection performance & Conceptualization/PoC & NA & NA & NA & NA & NA   \\
44 & \citet{paper_099} & WWW & 2023 & GPT-3, PaLM & Google Research & Finance & Sentiment Analysis & \begin{tabular}[c]{@{}l@{}}Labels generation for financial data\end{tabular} & FiQA-News & Fewshot (In-context learning) & Accuracy & Conceptualization/PoC & NA & NA & No & Yes & -   \\
45 & \citet{paper_010} & EMNLP Industry Track & 2023 & HyperCLOVA (30B and 82B), and GPT-3 & NAVER AI Lab & Generic & Societal Impact & Social bias risk mitigation& KoSBi & NA & F1 Score & Conceptualization/PoC & {\ul \href{https://github.com/naver-ai/korean-safety-benchmarks}{Link}} & Yes & Yes & Yes & \begin{tabular}[c]{@{}l@{}}The performance of the filter models are not very competitive\end{tabular}   \\
46 & \citet{paper_026} & EMNLP Industry Track & 2023 & \begin{tabular}[c]{@{}l@{}}GPT-3.5-turbo-0301, Falcon 40B-instruct, \\ Falcon 7B-instruct, Dolly-v2-12b\end{tabular} & Apple & Security & Societal Impact & \begin{tabular}[c]{@{}l@{}}Comprehensive handling of controversial issues\end{tabular} & DELPHI & Zero-shot & \begin{tabular}[c]{@{}l@{}}Controversy Acknowledgement Rate, \\ Comprehensiveness Answer Rate\end{tabular} & Conceptualization/PoC & {\ul \href{https://github.com/apple/ml-delphi}{Link}} & Yes & Yes & Yes & \begin{tabular}[c]{@{}l@{}}Dataset may not cover all the controversial questions.\\ and may contain expired ground truth controversy labels\end{tabular}   \\
47 & \citet{paper_106} & ICLR & 2023 & Decoder-only transformer model & AWS AI Labs & Software & Code generation & \begin{tabular}[c]{@{}l@{}} Evaluation of LLMs on multilingual \\ programming datasets \end{tabular} & MBXP, Multilingual HumanEval, MathQA-X & Zero-shot and Few-shot & pass@k scores & Conceptualization/PoC & {\ul \href{https://github.com/amazon-science/mxeval}{Link}} & Yes & NA & NA & \begin{tabular}[c]{@{}l@{}}Does not support language-specific functionalities\end{tabular}   \\
48 & \citet{paper_004}  & EMNLP Industry Track & 2023 & \begin{tabular}[c]{@{}l@{}}GPT-4, GPT3.5, PaLM-2, \\ LLaMA-2 13b, 7b\end{tabular} & Dialpad Canada Inc & NLP & Summarization & Business meeting summarization & QMSUM, AMI, ICSI & Zero-shot & ROUGE, BERTScore & Conceptualization/PoC & NA & Yes & Yes & Yes & \begin{tabular}[c]{@{}l@{}}Generalizability to domain-specific datasets is in question \\because only academic datasets were used for testing\end{tabular}   \\
49 & \citet{paper_018} & EMNLP  Industry Track & 2023 & FLAT-T5 & Amazon & NLP & Summarization & Summarization of length product titles & NA & NA & ROUGE, BLEU & Conceptualization/PoC & NA & NA & NA & NA & Do not guarantee inclusion of salient words in the summary \\
50 & \citet{paper_021} & RecSys & 2023 & Alpaca-LoRa & Sony Research India & Retail & Summarization & \begin{tabular}[c]{@{}l@{}}Generation of product descriptions sans webscrapping\end{tabular} & MovieLens, Goodreads Book graph & Fewshot (In-context learning) & \begin{tabular}[c]{@{}l@{}}Hit Rate, Normalized Discount \\ Cumulative Gain (NDCG), \\ Mean Reciprocal Rank (MRR)\end{tabular} & Conceptualization/PoC & NA & NA & NA & NA & \begin{tabular}[c]{@{}l@{}}Generates factually incorrect descriptions\end{tabular}   \\
51 & \citet{paper_098} & ESEC/FSE & 2023 & GPT-3.X & Microsoft & Infrastructure & Summarization & \begin{tabular}[c]{@{}l@{}}Cloud outage management\end{tabular} & Proprietary & NA & BLEU-4, ROUGE-L and METEOR & Deployment & NA & NA & NA & Yes & Evaluation metrics not fully reflect readability and usefulness of outage summary   \\
52 & \citet{paper_103} & CHI In2Writing Workshop & 2023 & NA & Allen Institute of AI & NLP & Summarization & \begin{tabular}[c]{@{}l@{}}Evidence based knowledge generation\end{tabular} & NA & Fewshot (In-context learning) & NA & NA & NA & NA & NA & Yes & NA   \\
53 & \citet{paper_002} & EMNLP Industry Track & 2023 & \begin{tabular}[c]{@{}l@{}}GPT4, TULU, Pythia,  Alpaca,\\  Vicuna, LLaMA-2, GPT-3.5\end{tabular} & Allen Institute for AI & NLP & Table-to-text-generation & \begin{tabular}[c]{@{}l@{}}Evaluation of LLMs on table-to-text generation\end{tabular} & LOTNLG, F2WTQ & Zero-shot and Few-shot & \begin{tabular}[c]{@{}l@{}}BLEU, ROUGE, SP-Acc, NLI-Acc, \\ TAPAS-Acc, TAPEX-Acc, Exact-match, \\ F1 Score, Accuracy\end{tabular} & Conceptualization/PoC & NA & Yes & NA & Yes & NA   \\
54 & \citet{paper_013} & ISWC & 2023 & GPT4, Llama2, FLAN-T5 & IBM Research & NLP & Table-to-text-generation & \begin{tabular}[c]{@{}l@{}}Automation of meta data\\ generation and enrichment\end{tabular} & NA & None & NA & Conceptualization/PoC & NA & NA & NA & NA & NA   \\
55 & \citet{paper_032} & NeurIPS & 2023 & GPT-3.5 (text-davinci-003 endpoint) & Microsoft & Generic & Table-to-text-generation &\begin{tabular}[c]{@{}l@{}} Noise induction to better \\understanding of table structures \end{tabular} & \begin{tabular}[c]{@{}l@{}}AirQuality, HousingData, Diabetes, \\ Wine Testing, Iris, Titanic, \\ and ENB2012\_data\end{tabular} & Zero-shot & F1 score & Conceptualization/PoC & {\ul \href{https://github.com/microsoft/prose}{Link}} & Yes & NA & NA & \begin{tabular}[c]{@{}l@{}}Performance of structural tasks with downstream task such as table \\ question answering remains an open challenge.\end{tabular}   \\
56 & \citet{paper_041} & IEEE Access & 2023 & ChatGPT & Microsoft & Robotics & Task Planning & \begin{tabular}[c]{@{}l@{}}Translating natural language instructions \\ to executable robot actions\end{tabular} & NA & Fewshot & Executability, Correctness & Conceptualization/PoC &  {\ul \href{https://github.com/microsoft/ChatGPT-Robot-Manipulation-Prompts}{Link}} & Yes & Yes & Yes & Only static environment is considered   \\
57 & \citet{paper_007} & EMNLP Industry Track & 2023 & GPT- text-embedding-ada-002 & Kingfisher Labs Ltd , Just Access & Legal & Tool & \begin{tabular}[c]{@{}l@{}}Automatic linkage of judgements to bookmarks \\ in court hearing videos\end{tabular} & UK National Archive & Zero-shot & Mean Average Precision (MAP), Recall & Conceptualization/PoC & NA & Yes & Yes & NA & NA   \\
58 & \citet{paper_050} & CHI Extended Abstract & 2023 & - & Google Research & Generic (HCI) & Tool & input-output interaction , Frame change & NA & Zero-shot & Questionnaire & Conceptualization/PoC & NA & NA & NA & Yes & \begin{tabular}[c]{@{}l@{}}Needs formal evaluation and in-depth analysis on \\how functional prompts effect prototyping process\end{tabular}   \\
59 & \citet{dibia-2023-lida} & ACL & 2023 & NA & Microsoft Research & Generic & Tool & \begin{tabular}[c]{@{}l@{}}Automatic generation of grammar-agnostic \\ visualizations and infographics\end{tabular} & Proprietary & Zero-shot and Few-shot & \begin{tabular}[c]{@{}l@{}}Visualization Error Rate (VER),\\ Self-Evaluated Visualization Quality\\ (SEVQ), code accuracy, \\ data transformation, goal compliance, \\ visualization type, data encoding,\\ and aesthetics\end{tabular} & Prototype & {\ul \href{https://microsoft.github.io/lida/}{Link}} & Yes & No & Yes & \begin{tabular}[c]{@{}l@{}}Code execution step increases computational complexity.\end{tabular}   \\
60 & \citet{paper_072} & ICRA & 2023 & \begin{tabular}[c]{@{}l@{}}text-davinci-*,\\ Codex,\\ GPT3\end{tabular} & Nvidia corporation & Robotics & Tool & Generate programmatic robot instructions using LLMs & NA & Fewshot (In-context learning) & \begin{tabular}[c]{@{}l@{}}Success rate(SR),\\ Goal condition recall(GCR),\\ Executability(Exec)\end{tabular} & Development & {\ul \href{https://github.com/NVlabs/progprompt-vh}{Link}} & Yes & No & Yes & \begin{tabular}[c]{@{}l@{}} Robotic action success feedback is not\\ shared with agent leading to failure scenarios \end{tabular}   \\
61 & \citet{paper_075} & EACL & 2023 & \begin{tabular}[c]{@{}l@{}}mT5-Large,\\ PaLM\end{tabular} & Google research India & NLP & Translation & Translating English datasets to several other languages & MTOP, MASSIVE & Zero-shot and Few-shot & corrected Exact Match & Prototype & NA & NA & No & Yes & \begin{tabular}[c]{@{}l@{}} Computationally expensive\end{tabular}   \\
62 & \citet{paper_016} & ACL & 2023 & BERT, ALBERT, RoBERTa & IBM Research & Generic & Trustworthy AI & Inspect the fairness and Bias of foundation models & CrowS-Pairs & NA & NA & Conceptualization/PoC &  {\ul \href{https://github.com/IBM/finspector}{Link}} & NA & NA & Yes & \begin{tabular}[c]{@{}l@{}}The tool's effectiveness not tested for decoder-only models.\end{tabular}   \\
63 & \citet{paper_019} & EMNLP & 2023 & \begin{tabular}[c]{@{}l@{}}GPTNeo-2.7B, GPTJ-6B, \\ Open-LLaMA-7B, \\ RedPajama-7B,GPT3.5-Turbo, \\ GPT4\end{tabular} & Amazon Alexa AI & Security & Trustworthy AI & Evaluating the LLMs for Hallucinations & DBpedia, TriviaQA & Zero-shot & \begin{tabular}[c]{@{}l@{}}BLEU, ROUGE, METEOR, \\ BERTScore, AlignScore\end{tabular} & Conceptualization/PoC &  {\ul \href{https://github.com/amazon-science/invite-llm-hallucinations}{Link}} & NA & NA & NA & \begin{tabular}[c]{@{}l@{}}Lack of diversity in test set.\end{tabular}   \\
64 & \citet{paper_027} & CI & 2023 & Not disclosed & Apple & Generic & Trustworthy AI & Identify the presence of gender bias in LLMs & Proprietary & Zero-shot & NA & Conceptualization/PoC & NA & NA & NA & Yes & Results may not reflect the real gender bias   \\
65 & \citet{paper_028} & EMNLP & 2023 & ChatGPT, Alpaca & Adobe Research & Generic & Trustworthy AI & Identify the presence of gender bias in LLMs & WikiBias-Aug & Zero-shot & WEAT & Conceptualization/PoC &  {\ul \href{https://github.com/uclanlp/biases-llm-reference-letters}{Link}} & NA & Yes & Yes & \begin{tabular}[c]{@{}l@{}}Only consider binary gender when analyzing biases\end{tabular}   \\
66 & \citet{paper_056}  & NeurIPS & 2023 & \begin{tabular}[c]{@{}l@{}}OPT-350M\\ OPT-1.3B\\ OPT- 2.7B\end{tabular} & NAVER AI Lab, Parameter Lab & Generic & Trustworthy AI & Probing for PII in a given LLM & Pile & Fewshot (In-context learning) & \begin{tabular}[c]{@{}l@{}}Sentence matching,\\ Likelihood ratio\end{tabular} & Conceptualization/PoC & NA & NA & Yes/ & Yes & \begin{tabular}[c]{@{}l@{}}Evaluation dataset uses private\\information sourced from open-source datasets\end{tabular}   \\
67 & \citet{paper_066} & \begin{tabular}[c]{@{}l@{}}EMNLP System \\ demonstrations\end{tabular} & 2023 & \begin{tabular}[c]{@{}l@{}}text-davinci-003, GPT-3.5-turbo, \\ falcon-7b-instruct, llama2-13b-chat\end{tabular} & NVIDIA & Generic & Trustworthy AI & \begin{tabular}[c]{@{}l@{}}Tool kit for adding programmable guardrails for \\ conversational LLMs\end{tabular} & \begin{tabular}[c]{@{}l@{}}Anthropic Red-Teaming and\\ Helpful datasets\end{tabular} & Fewshot (In-context learning) & Accuracy & Development & {\ul \href{https://github.com/NVIDIA/NeMo-Guardrails/}{Link}} & Yes & NA & Yes & \begin{tabular}[c]{@{}l@{}}Toolkit not suitable as standalone solution\end{tabular}   \\
68 & \citet{paper_069} & \begin{tabular}[c]{@{}l@{}}EMNLP System \\ demonstrations\end{tabular} & 2023 & Generic & {\ul H2O.ai} & Generic & Trustworthy AI & \begin{tabular}[c]{@{}l@{}}Deploy and test efficiency of wide variety of LLMs \\ on private databases and documents\end{tabular} & NA & NA & NA & Deployment &  {\ul \href{https://github.com/h2oai/h2ogpt}{Link}} & Yes & Yes & Yes & \begin{tabular}[c]{@{}l@{}}Datasets, Biases and Offensiveness, Usage, Carbon footprint, \\ Hallucinations of LLMs\end{tabular}  \\
\bottomrule
\end{tabular}%
}
\caption{Master table of the survey with 68 research papers.}
\label{tab:check_list}
\end{sidewaystable}
\end{document}